\RequirePackage{fix-cm}

\documentclass[twocolumn]{svjour3}          
\smartqed  
\usepackage{graphicx}
\usepackage{amsmath}
\usepackage{color}
\usepackage{multirow}
\usepackage{subfigure}
\usepackage{array}
\usepackage[backref]{hyperref} 
\usepackage{float}
\usepackage[keeplastbox]{flushend}
\usepackage{url}
\usepackage{ulem}
\usepackage{pifont}
\usepackage{xspace}
\newcommand{\ie}{\textit{i.e.}\xspace}
\newcommand{\eg}{\textit{e.g.}\xspace}

\newcommand{\wrt}{\textit{w.r.t.}\xspace}

\newcommand{\resp}{\textit{resp.}\xspace}

\usepackage{geometry}
\begin{document}

\title{AutoScale: Learning to Scale for Crowd Counting
}


\author{Chenfeng Xu \thanks{\textit{Disclaimer}: This work was mostly done when Chenfeng Xu was an undergraduate student in Huazhong University of Science and Technology.}* \and Dingkang Liang* \and Yongchao Xu \and Song Bai \and Wei Zhan  \and Xiang Bai \and Masayoshi Tomizuka
}


\institute{* Equal contribution
\\\
Chenfeng Xu, Wei Zhan, and Masayoshi Tomizuka are with University of California, Berkeley.
E-mail: (xuchenfeng, wzhan, tomizuka)@berkeley.edu
\\\\
Dingkang Liang and Xiang Bai are with Huazhong University of Science and Technology.
E-mail: (dkliang, xbai)@hust.edu.cn
\\\\
        Yongchao Xu is with Wuhan University (corresponding author).
        E-mail: yongchao.xu@whu.edu.cn
        \\\\
        Song Bai is with ByteDance and University of Oxford, UK. 
        E-mail: songbai.site@gmail.com
}

\date{Received: date / Accepted: date}

\maketitle

\begin{abstract}
Recent works on crowd counting mainly leverage Convolutional Neural Networks (CNNs) to count by regressing density maps, and have achieved great progress. In the density map, each person is represented by a Gaussian blob, and the final count is obtained from the integration of the whole map. However, it is difficult to accurately predict the density map on dense regions. A major issue is that the density map on dense regions usually accumulates density values from a number of nearby Gaussian blobs, yielding different large density values on a small set of pixels. This makes the density map present variant patterns with significant pattern shifts and brings a long-tailed distribution of pixel-wise density values.
In this paper, we aim to address such issue in the density map. Specifically, we propose a simple and effective Learning to Scale (L2S) module, which automatically scales dense regions into reasonable closeness levels (reflecting image-plane distance between neighboring people). L2S directly normalizes the closeness in different patches such that it dynamically separates the overlapped blobs, decomposes the accumulated values in the ground-truth density map, and thus alleviates the pattern shifts and long-tailed distribution of density values. This helps the model to better learn the density map. We also explore the effectiveness of L2S in localizing people by finding the local minima of the quantized distance (\textit{w.r.t.} person location map), which has a similar issue as density map regression. To the best of our knowledge, such localization method is also novel in localization-based crowd counting. We further introduce a customized dynamic cross-entropy loss, significantly improving the localization-based model optimization.
Extensive experiments demonstrate that the proposed framework termed \textit{AutoScale} improves upon some state-of-the-art methods in both regression and localization benchmarks on three crowded datasets and achieves very competitive performance on two sparse datasets. An implementation of our method is available at \url{https://github.com/dk-liang/AutoScale.git}.

\end{abstract}

\begin{figure*}[!tb]
\centering
\subfigure[Ground-truth density map representation (middle) and predicted density map (right) with AutoScale.]
 {
 \begin{minipage}[tb]{0.9\textwidth}
 \centering
\includegraphics[width=14cm,height=4cm]{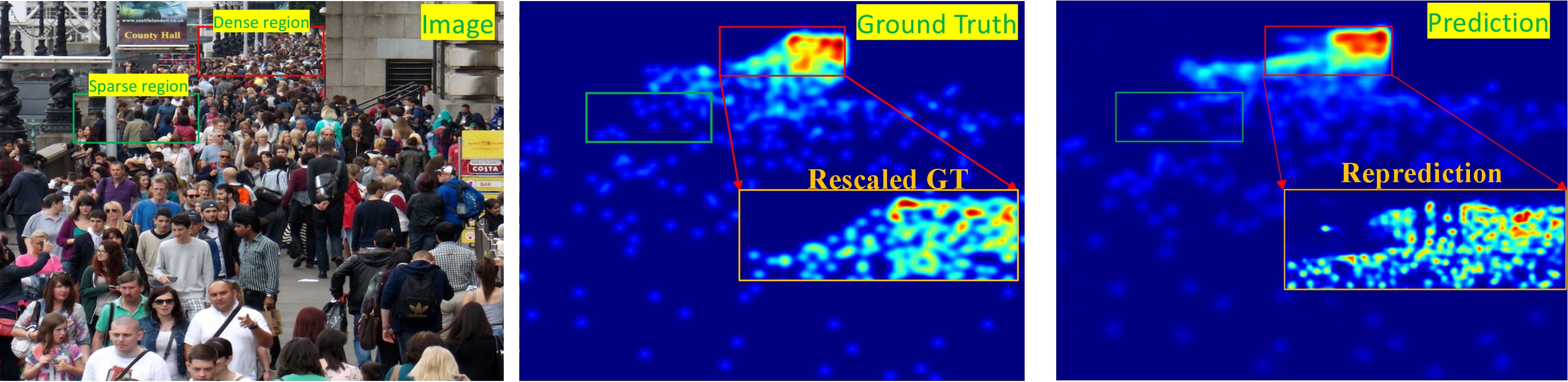}
 \end{minipage}
 \label{fig:intro_1}
 }
 \subfigure[Distribution of pixel values in ground-truth (left) and predicted (right) density map on the selected regions.]
 {
 \begin{minipage}[tb]{0.9\textwidth}
 \centering
\includegraphics[width=14cm,height=4cm]{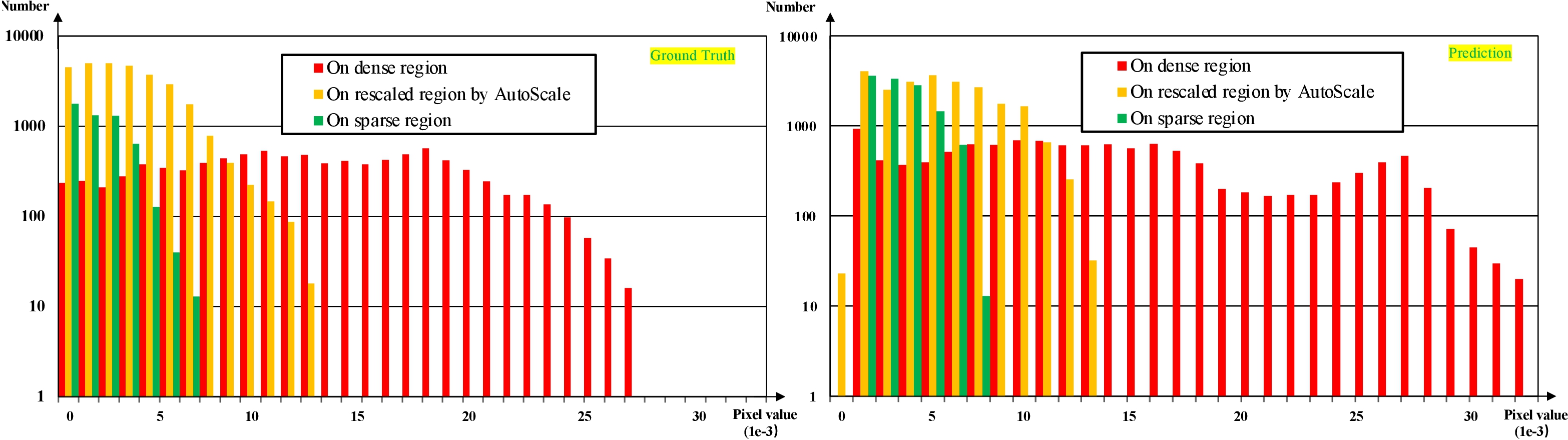}
 \end{minipage}
 \label{fig:intro_2}
 }
\caption{The intuitive and statistical distribution of the density map. (a) Intuitively, we can observe that the Gaussian blobs distribute separately and are similar to each other in the sparse region (\eg, within the green box), while variant overlaps exist in the dense region (\eg, within the red box). Re-scaling the dense region via closeness normalization alleviates the pattern shifts between dense and sparse regions. (b) Statistically, the dense region (\eg, in the red box) presents a long-tailed distribution of density values. The proposed AutoScale alleviates this issue and reduces the density value distribution gap between dense and sparse regions, thus facilitating density map regression on dense regions. Note that the rescaled GT is obtained using the predicted scale factor and the Y-axis of (b) is in logarithmic scale}
\label{fig:distribution map}
\end{figure*}

\section{Introduction}
\label{intro}
Crowd counting has recently attracted great interest owing to its importance in a wide range of applications, \textit{e.g.}, video monitor~\cite{kang2018beyond}, public security and city management~\cite{sindagi2018survey}. Although the \textit{de facto} CNN-based methods~\cite{zhao2016crossing,zhang2016single,shi2019revisiting,liu2019context,sam2017switching} have made significant progresses over traditional methods~\cite{chan2008privacy,chen2012feature,ge2009marked,idrees2015detecting}, it is still very difficult to accurately reason the count especially in dense regions where the crowd gathers. Whereas, the very crowded scene full with people is very common in real life, such as large gathering, train station, stadium.

In particular, most methods aim to estimate accurate and high-quality density maps that represent people through Gaussian blobs and can be integrated to obtain the final counts. However, there exist multiple Gaussian blob overlaps in dense regions, thus the density value of one pixel can be accumulated from many different nearby Gaussian blobs. Meanwhile, these accumulated density values are usually quite crucial to the final count yet hard to accurately predict. 
For instance, as shown in Fig.~\ref{fig:intro_1}, we can observe that the Gaussian blobs in the sparse region are separated and similar to each other, while gather and overlap in the dense region. Statistically, as shown in Fig.~\ref{fig:intro_2}, different from the distribution of density values in the sparse region, it presents a long-tailed shape in the dense region, which hinders the model from accurate density map prediction for the following reasons:

1) \textbf{Open end of density values:} The density values on dense regions are usually accumulated in multiple ways from nearby Gaussian blobs, leading to an open end of the density values in dense regions.

2) \textbf{Density value imbalance:} Pixels in the dense region usually have large density values and only occupy a small part of the whole density map, which poses density value imbalance. However, the count on dense regions is crucial for accurate crowd counting. 

3) \textbf{Density distribution gap:} As show in Fig.~\ref{fig:intro_2}, there exists a huge density distribution gap between the sparse and dense regions. Besides, the density distributions of dense regions from different images also vary a lot because of variant value accumulations. This further gives rise to pattern shifts.

In fact, both the
pattern shift and long-tailed distribution bring challenge for accurate prediction in dense regions.
In this paper, to the best of our knowledge, we are the first to try to mitigate such issue in density map for crowd counting. To this end, we propose a simple yet effective Learning to Scale (L2S) module to automatically learn reasonable scale factors, and then rescale the dense regions into similar and appropriate closeness levels reflecting image-plane distance between neighboring people. This helps to separate the overlapped blobs and decompose the original accumulated density values in density map. Therefore, normalizing the closeness alleviates the issue of pattern shift and long-tailed distribution by pattern normalization, hence facilitates the regression of density map.
It is noteworthy to mention that since there is no ground-truth for the scale factor suggesting how much a given dense region should be zoomed ideally, the proposed L2S performs in an unsupervised clustering way via the center loss.
An example showing the effect of the proposed L2S is illustrated in Fig.~\ref{fig:intro_1} and Fig.~\ref{fig:intro_2}. It can be observed that through L2S, the long-tailed density distribution and pattern shift are mitigated.

The localization-based method is currently attracting much attention recently. Instead of detecting each person, we propose to regard the local minima of the quantized distance (\wrt person location) map as the final person localization result. Specifically,
we term such quantized distance map as \textit{distance-label map}, which divides different distances into a number of categories, representing different distance ranges. The local minima of such distance-label map correspond to the localization of people. To the best of our knowledge, we are the first to leverage the local minima of such distance-label map for localization-based crowd counting. Similar to the density map representation, there exist class imbalance and distribution variances. This motivates us to employ L2S on distance-label map to separate the closed blobs and mitigate the distribution variances to improve the localization accuracy. Besides, we also design a customized dynamic cross-entropy (DCE) loss to guide the distance-label map learning. Specifically, different from the widely used static weighted cross-entropy loss, the weights are generated by the multiplication between the prediction possibilities and the absolute difference between predicted class and ground truth class, which is to dynamically change according to the prediction. 

We leverage a FPN-like baseline model to frame both of our regression-based method and localization-based method with the proposed L2S module, which is termed \textit{AutoScale}. Precisely, the baseline model provides an initial prediction, giving count for sparse regions and helping to automatically select a dense region for further refinement. The proposed L2S module generates an appropriate scale factor to rescale the selected dense region. A second count uses the same FPN-like model is performed on the rescaled dense region, and replaces the initial count on that region. We adopt the similar pipeline to the localization-based method by simply changing the output from the density map to the distance-label map.

Extensive experiments demonstrate that AutoScale outperforms some state-of-the-art methods on UCF-QNRF, JHU-Crowd++, NWPU-Crowd datasets for both regression-based and localization-based methods, and achieves very competitive performance on ShanghaiTech Part A and ShanghaiTech Part B dataset. On the extracted dense regions, where the accurate counting is very challenging, the proposed L2S achieves significant improvements. Besides, applying the L2S to some other popular crowd counting methods consistently improves their corresponding performance. Moreover, we also conduct the experiments on the TRANCOS dataset, further showing the superiority of the distance-label map with customized dynamic cross-entropy loss for localization-based method. 

The main contribution of this paper lie in three folds: 1) we are the first to explore the long-tailed distribution of pixel-wise values in density map for crowd counting and propose the Learning to Scale (L2S) module to mitigate this issue. 2) We are -to the best of our knowledge- the first to localize people by finding the local minima of distance-label map in localization-based crowd counting. We also propose a novel customized dynamic cross-entropy loss, which takes advantage of the geometrical meaning of distance-label map. This mitigates the class imbalance and significantly improves the baseline performance of the localization-based method. 3) The proposed regression-based AutoScale (\textit{resp.} localization-based AutoScale) based on a simple baseline consistently outperforms some regression-based (\textit{resp.} localization-based) state-of-the-art methods on three public dense datasets and achieves very competitive performance on two sparse datasets. Besides, the proposed L2S is also helpful in improving the performance of some other popular methods.


The current paper extends the preliminary study of this work~\cite{xu2019learn} in the following four major aspects:
\begin{itemize}
    \item First, we reformulate the intuition from the point of view of long-tailed distribution, which reveals better the mechanism of L2S in improving the counting accuracy. To the best of our knowledge, L2S is the first attempt that explicitly explores the pixel-level long-tailed distribution issue in density map regression and localization-based crowd counting. 
    
    \item Second, in addition to density map regression, we also adapt the proposed L2S to localization-based crowd counting. Specifically, we novelly leverage the distance-label map to count by localizing the local minima, and design the customized dynamic cross-entropy (DCE) loss. Sufficient experiments have been conducted to demonstrate that the proposed L2S is also effective on the localization-based method. 
    
    \item  Third, we improve the original method in the conference version~\cite{xu2019learn} via some more reasonable designs (\eg, adaptive dense area selection instead of regular patches given by evenly dividing the image domain and the distance-based density measurement). The average distance between nearby persons is a more intuitive and efficient indication about how dense the crowd is.  
    
    \item Last, we conduct many more comparison experiments on more challenging datasets to demonstrate the effectiveness of the proposed L2S and DCE loss. We also now deeply analyze the effectiveness of L2S on dense regions and in combination with different baseline methods.
\end{itemize}

\section{{Related Work}}
 \label{sec:relatedwork}
We shortly review some related works on crowd counting in Section~\ref{subsec:cc}, other vision tasks involving scaling operation in Section~\ref{subsec:scaling} and long-tailed distribution problems in Section~\ref{subsec:long-tailed}.

\subsection{Crowd counting}
\label{subsec:cc}
Current mainstreams for crowd counting consist of two kinds of methods, localization-based methods and regression-based methods. We shortly review some representative works in the following.

\noindent\textbf{Regression-based methods.} 
Regression-based methods are existing mainstreams in crowd counting, thanks to the widely used density map. Before the era of deep learning, previous works~\cite{chan2008privacy,chen2012feature,ge2009marked,idrees2015detecting,liu2015bayesian,arteta2014interactive} resort to different regression strategies,~\textit{e.g.}, linear regression, Gaussian regression, ridge regression. Current regression-based methods~\cite{zhang2016single,sam2017switching,cao2018scale,li2018csrnet,sindagi2017generating,wang2019learning,wan2019residual,liu2019adcrowdnet,zhao2019leveraging,zhao2019leveraging,lian2019density,jiang2019learning,shi2018crowd,zhang2019nonlinear} leverage CNNs to regress density maps, based on which not only the count but also the approximate distribution can be reasoned.

Though density map regression-based methods have achieved significant progress, there still exist several challenges, such as scale variations, perspective distortions, and noisy background interference. Multi-scale feature fusion~\cite{zhang2016single,cao2018scale,onoro2016towards,jiang2019crowd,sindagi2019multi,sindagi2020jhu,sindagi2017generating,ranjan2018iterative,zhang2019wide,liu2020adaptive} is an effective way to improve the ability of coping with different scales.
AMSNet~\cite{hu2020count} relies on network architecture search (NAS) to automatically adopt a multi-scale architecture, addressing the scale variation issue in crowd counting.
 Some works~\cite{kang2018crowd,sajid2020zoomcount} aim to scale the image/attended region according to fixed scale factors.
The strategy of dividing and conquering is also applied to crowd counting. Sam~\textit{et al.}~\cite{sam2017switching} and Babu~\textit{et al.}~\cite{babu2018divide} adopt the scale classifiers~\cite{sam2017switching,babu2018divide} to predict the scale levels of different regions and further design models to separately deal with them. Liu~\textit{et al.}~\cite{liu2018decidenet} attempt to count people in sparse regions through detection methods and regress density maps in dense regions. S-DCNet~\cite{xiong2019open} transforms the open-set problem into the close-set problem by dividing spatial planes on the feature map. Attention mechanism is another trend of to cope with spatial relations in crowd counting, such as self-attention (non-local) module~\cite{zhang2019relational,wan2019adaptive} and other customized attention blocks~\cite{zhang2019attentional,miaoshallow,hossain2019crowd,liu2018decidenet,liu2018crowd,miaoshallow}. 

Some auxiliary tasks are widely combined to improve density map estimation,~\textit{e.g.}, foreground and background segmentation ~\cite{liu2019adcrowdnet,zhao2019leveraging,jiang2019mask,shi2019counting}, depth predictions~\cite{zhao2019leveraging,lian2019density}, crowd velocity estimations~\cite{zhao2016crossing}, uncertainty estimation~\cite{oh2019crowd}, and target correction~\cite{bai2020adaptive}. The foreground masks~\cite{liu2019adcrowdnet,zhao2019leveraging,shi2019counting} or trimap~\cite{arteta2016counting} given by thresholding the distance map are usually used for filtering out noisy predictions in the background so that the estimation biases are relatively removed and the predictions become more accurate. The depth estimations~\cite{zhao2019leveraging,lian2019density} provide information about people scales, which are beneficial for density map estimation. Liu~\textit{et al.}~\cite{liu2018leveraging,liu2019exploiting} leverage the multi-task (learn to count and learn to rank) strategy to train the simple baseline~\cite{vgg16network} and effectively facilitate the estimations of density maps. CAN~\cite{liu2019context}, PACNN~\cite{shi2019revisiting}, 3DCC~\cite{zhang20203d}, PRNet~\cite{yang2020reverse}, and PGCNet~\cite{yan2019perspective} extract perspective knowledge to help CNNs adapt to diverse scales. 
ASNet~\cite{jiang2020attention} learns several extra masks of multiplication rates to automatically adjust the density estimation of each corresponding sub-region. 
HyGnn~\cite{luo2020hybrid} leverages hybrid graph neural network to learn localization map as the auxiliary task to enhance density map prediction for crowd counting.

Besides the above model design mechanisms, the objective function is also an important direction. In particular, Bayesian loss~\cite{ma2019bayesian} is proposed to regard the density map as a probability map and compute the probability of each pixel. Cheng \textit{et al.}~\cite{cheng2019learning} proposed Maximum Excess over Pixels (MEP) loss, which finds pixel-level region with high difference to the ground truth, and then the region is selected for optimization. DSSINet~\cite{liu2019crowd} utilizes a Dilated Multiscale Structural Similarity (DMSSSIM) loss to produce locally consistent density maps. 

\noindent\textbf{Localization-based methods.} Traditional methods for crowd counting try to count by detecting person faces~\cite{chen2010people,zhao2009people} or directly detecting pedestrians~\cite{wang2011automatic,viola2005detecting,brostow2006unsupervised,rodriguez2011density}. Nevertheless, box annotations for detection are quite laborious especially in extreme dense regions. A compromised kind of annotation is to point out the exact location (\textit{e.g.}, center point of head) of each person. Such annotation in terms of individual points hinders the use of powerful object detection pipelines~\cite{girshick2015fast,ren2015faster,he2017mask}. Besides, detection methods usually suffer from severe occlusions in highly congested regions. Despite these difficulties, detection/localization methods have witnessed great progresses. Laradji~\textit{et al.}~\cite{laradji2018blobs} proposes a watershed split loss based on the distance map for separating nearby people. Ribera~\textit{et al.}~\cite{ribera2019} propose a loss based on weighted Hausdorff distance (WHD) and formulate the object-localization problem as the minimization of distances between points.
Liu~\textit{et al.}~\cite{liu2019recurrent} observe that zooming in (at a fixed rate) the dense region is effective for further localization. To tackle the shortage of the box annotations, Liu~\textit{et al.}~\cite{liu2019point} 
and Sam~\textit{et al.}~\cite{sam2020locate} propose an impressive method to detect bounding boxes under the supervision of point-level annotations. Idrees~\textit{et al.}~\cite{idrees2018composition} attempt to localize people based on local maxima of predicted density map with a small Gaussian kernel. 

Our work is different from the above methods. We explore how to learn better density maps from the aspect of distribution of density values. We propose a novel L2S module to effectively alleviate the long-tailed density distribution in dense regions, helping to better regress density maps. We also demonstrate the effectiveness of proposed method in localizing people in dense regions by regarding the local minima of distance-label map as person localization result.


\subsection{Scaling in vision tasks}
\label{subsec:scaling}
Scaling plays an important role in many vision tasks, \textit{e.g.}, object detection~\cite{singh2018analysis,singh2018sniper,najibi2018autofocus} and fine-grained classification~\cite{zheng2017learning,recasens2018learning}. Singh~\textit{et al.}~\cite{singh2018analysis} propose a method called scale normalization of image pyramid by selecting objects with relative similar scales. Instead of processing an entire image pyramid, SINPER~\cite{singh2018sniper} processes context regions around ground-truth instances at the appropriate scale. Najibi~\textit{et al.}~\cite{najibi2018autofocus} attempt to design an efficient algorithm to automatically focus on small objects that are usually hard to detect, then process them at finer scales. In the field of fine-grained classification, zooming in attended regions is an effective method to better recognize specific objects. For example, Zheng~\textit{et al.}~\cite{zheng2017learning} propose an attention method to search for key regions with important features for specific fine-grained classes and zoom in the regions to see better. Similar to~\cite{zheng2017learning}, Recasens~\textit{et al.}~\cite{recasens2018learning} also attempt to find salient regions and zoom in them for better fine-grained object classification. It is noteworthy to mention that STN~\cite{jaderberg2015spatial} also changes the original scales by learning parameters of affine transformation without specific supervision. 

Most existing methods~\cite{recasens2018learning,jaderberg2015spatial,liu2019recurrent,sajid2020zoomcount,kang2018crowd,singh2018analysis,singh2018sniper} involving scaling operations mainly implicitly find attended or important regions and rescale them to regions of fixed size (or at a fixed zooming rate) for better detection or recognition. 
The proposed AutoScale differs from those existing methods in scaling purpose, motivation, and computation of scale factor. Typically, many works~\cite{liu2019recurrent,sajid2020zoomcount,kang2018crowd,singh2018analysis,singh2018sniper} aim to scale the image according to fixed scale factors in order to have a more accurate attention. AutoScale aims to explicitly select dense regions and then rescale them into similar closeness levels, thus alleviating the long-tailed issues in dense regions.

\begin{figure*}[htb!]
\centering
\includegraphics[width=0.8\paperwidth]{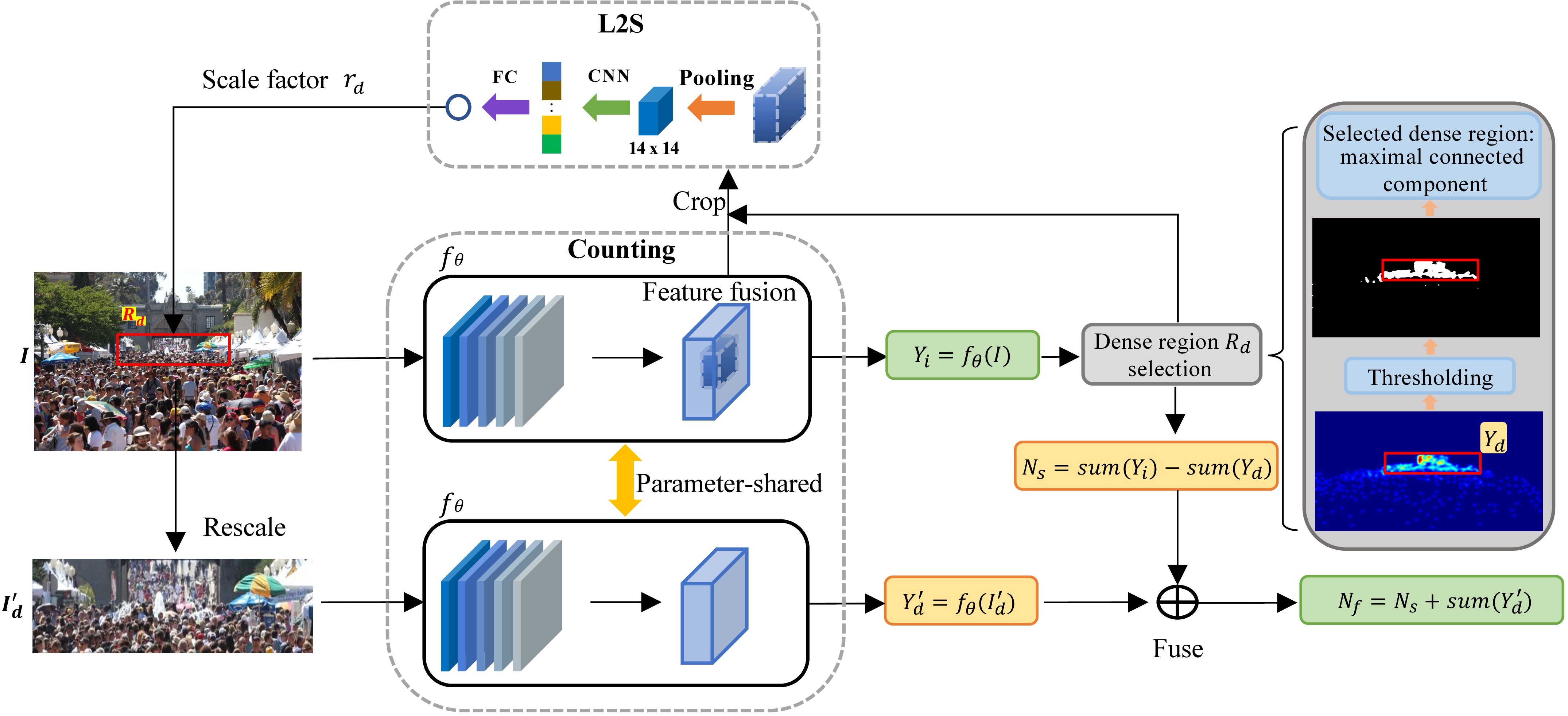}
\centering
\caption{The diagram of the proposed AutoScale. There are two major modules: 1) the basic counting module $f_\theta$ dedicated for both initial prediction $Y_i$ on original image $I$ and re-prediction $Y'_d$ on the rescaled dense region $I'_d$ for both regression-based and localization-based counting; 2) the L2S module that generates an appropriate scale factor $r_d$ to rescale the selected dense region $R_d$ for refining count on the dense region $R_d$. The final count $N_f$ is composed of initial count on sparse regions $N_s = sum(Y_i) - sum(Y_d)$, where $Y_d$ is the initial prediction on the selected dense region $R_d$, and re-predicted count on the selected dense region $sum(Y'_d)$}.
\label{fig:pipeline}
\end{figure*}

\subsection{Long-tailed distribution in vision tasks}
\label{subsec:long-tailed}
Long-tailed distribution is extremely common in natural data, and has been extensively studied ~\cite{zhu2014capturing,salakhutdinov2011learning,zhu2016we,ouyang2016factors,van2017devil,cui2018large,he2009learning,geng2020recent}. It is very classical~\cite{he2009learning} to mitigate the long-tailed distribution by under-sampling the major classes, over-sampling the minor classes and re-weighting the data. Most recent works mainly focus on improving objective functions~\cite{lin2017focal,zhang2017range,oh2016deep,cao2019learning}, training strategies~\cite{dong2017class} and model design~\cite{wang2017learning,ha2016hypernetworks}. The improvement by studying the long-tailed problem mainly exist in the current mainstreaming vision applications, such as recognition, object detection and segmentation. 

There still exist severe long-tailed distribution issue yet rarely explored in crowd counting, especially for the aspect of density map. The most related work to our proposed method is S-DCNet~\cite{xiong2019open}, which also aims to tackle the open-set problem in crowd counting by dividing the spatial planes to maintain the count under an closed range. 
Although S-DCNet~\cite{xiong2019open} also explores the long-tailed issue in crowd counting, it is quite different with our AutoScale. In fact, we explore the pixel-level long tailed distribution of density values caused by overlapped Gaussian blobs on the highly congested regions of the density map. Whereas, S-DCNet~\cite{xiong2019open} addresses the long-tailed distribution of image/patch-level people number. Besides,
dividing the spatial planes is hardly to cope with the long-tailed distribution resulting from the accumulations of pixel value. Different from this, we propose L2S to separate the overlapped blobs and decompose the original accumulated values, which is dynamical and with less hand-craft operations. Furthermore, the proposed method is also beneficial to improve the localization accuracy for methods based on distance 
label maps.

\section{Learn to scale for crowd counting}
\label{sec:method}

\subsection{Overview}
\label{subsec:overview}

The long-tailed distribution in density map poses great challenges to crowd counting, yet few works focus on mitigating it. Specifically, the accumulated density values in dense regions result in the open end, value imbalance and huge distribution gap between dense and sparse regions. Nevertheless, these density values in dense regions are quite crucial to the final prediction. 

Consequently, we propose a learning to scale (L2S) module, acting as an unsupervised clustering that leverages the center loss to rescale all dense regions into similar and reasonable closeness levels, which mitigates the open end, transforms the distributions into as similar as sparse regions and reduces the distribution gap between dense regions. We build the framework using a FPN-like baseline network with L2S to demonstrate the effectiveness of the proposed method. We first apply such L2S module into the baseline model that regresses density map and demonstrates the effectiveness of the proposed L2S. Then we simply change the output into distance-label map that can be used for counting by localization, which further demonstrates the effectiveness of L2S on localization-based method. Besides, we propose a novel dynamic cross-entropy loss for distance-label map to better learn the distance-label map representation, further boosting the localization accuracy. Note that we provide two kinds of methods, regression-based AutoScale and Localization-based AutoScale, which share similar networks but are independent with each other and used for counting by regression and counting by localization, respectively.

Both the regression and localization-based AutoScale with L2S are end-to-end trainable. The overall framework is depicted in Fig.~\ref{fig:pipeline}. The whole pipeline consists of two parts: 1) Counting network based on a widely used backbone (\textit{e.g.}, FPN~\cite{lin2017feature} in this paper) for the estimations of density maps and distance-label maps; 2) L2S dedicated for generating appropriate scale factors for selected dense regions. We will detail the proposed methods in the following.

\begin{figure*}
\centering
\resizebox{0.97\textwidth}{!}{
    \includegraphics{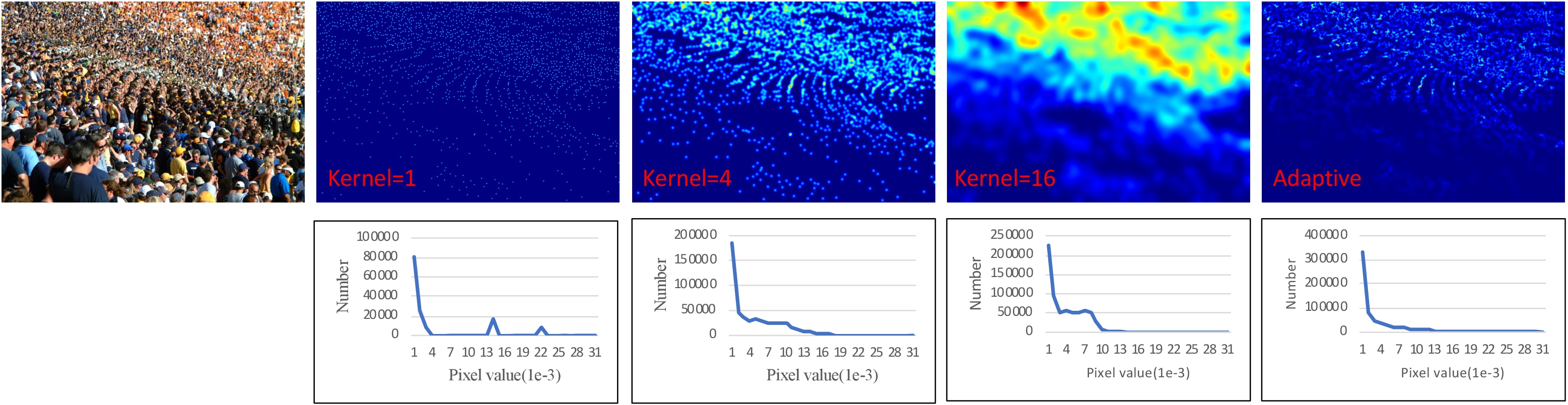}
    }
    \caption{The density maps with different Gaussian kernels and the corresponding distributions of density values.}
\centering
\label{fig:different_kernel}
\end{figure*}
\subsection{Analysis on density map representation}
\label{subsec:adm}

Most current crowd counting methods rely on density map regression to count people. However, few works aim to explore the distribution of pixel values in density map. Generally, most recent crowd counting datasets provide point-level annotations, which can be represented as binary maps $B$. For each pixel in the image domain $p \in \Omega$, we have $B(p)=\sum_{i=1}^P \delta(p-p_i)$, where each head position $p_i$ is modeled as a delta function $\delta(p-p_i)$, and $P$ refers to the total number of people in the image.
The density map $D$ on each pixel $p$ is then generated by convolving $B(p)$ with a Gaussian kernel $G$~\cite{zhang2016single}: $D(p) = \sum_{i=1}^P{\delta(p - p_i)} * G$. Meanwhile, the kernel size of $G$ is a key hyper-parameter for crowd counting\cite{zhang2016single,wan2019adaptive}.

In particular, a larger kernel size is prone to bring in more overlaps, therefore the pixel accumulations occur more, while a smaller kernel size leads to more severe value imbalances. Although adaptive Gaussian kernel proposed in~\cite{zhang2016single} can reduce the overlaps of Gaussian blobs by setting the kernel size according to the K nearest neighborhood distance, the distances are quite variant, making an open end distribution. An example is presented in Fig.~\ref{fig:different_kernel}. Except for Gaussian kernel equal to 1, others present the long-tailed distribution because of multiple pixel accumulations from the overlaps of Gaussian blobs. Yet, as for the Gaussian kernel equal to 1, there exist severe pixel value imbalance, which is also difficult for CNNs to learn. Kernel generator~\cite{wan2019adaptive} is a promising method to cope with the dilemma of kernel selection. However, it is not explicit to deal with the distribution of the density values.

To the best of our knowledge, we are the first to explore how to better learn the density map from the aspect of pixel value distribution and propose the Learn to Scale (L2S) to dynamically change the distribution of density map in the dense regions, which makes the model be able to better learn and represent the density map during both training and inference phase.

\subsection{Learn to scale}
\label{subsec:l2s}

Learning to scale aims to modulate dense regions of different scales to similar and appropriate closeness levels, which tackles the problem of long-tailed distribution for density map. Precisely, for a given region $R$, we define a closeness level $S$ for $R$ via ground-truth as
\begin{equation}
S = \frac{\sum_{i=1}^{P_R} d_i}{P_R},
\label{eq:scalelevel}
\end{equation}
where $d_i$ denotes the distance between $i$-th person and its nearest person in $R$, and $P_R$ stands for the overall number of people in the region $R$. 
Note that different from the conference version~\cite{xu2019learn} that relies on the average people number to define the density level, we now leverage the average distance. Actually, dense (\resp, sparse) region intrinsically contains people that lie very close to each other (\resp, far away from each other). Therefore, the closeness level for both dense and sparse region could be naturally measured by the average distance between each person and its nearest person within the region. More importantly, different from patch-wise and pixel-wise crowd number, the average distance is not influenced by the area without people, the closeness level defined in Eq.~\eqref{eq:scalelevel} is thus more reliable and intuitive.

For a given dataset consisting of $N_R$ regions to be rescaled, we target scaling each region of interest $R_i \, (i = 1, 2, \dots, N_R)$ such that the closeness level of each rescaled region approaches a similar scale $\overline{S}$. For that, we need to generate an appropriate scale factor $r_i$ for each region $R_i$. However, there is no explicit target scale factor suggesting how much the region $R_i$ should be zoomed and also no target similar closeness level $\overline{S}$ indicating which should be approached. Therefore, we propose a learn to scale module that acts as an unsupervised clustering by using the center loss on closeness levels.

Specifically, for each region $R_i \, (i = 1, 2, \dots, N_R)$ to be rescaled, we attempt to generate a corresponding scale factor $r_i$. We apply a simple CNN consisting of three $3 \times 3$ convolutional layers and two fully connected layers on the backbone feature of region $R_i$ to produce $r_i$, which is then used to rescale the region $R_i$ via bilinear upsampling. Such L2S module is learned using the following training objective in terms of the center loss on closeness levels
\begin{equation}
L_s \, = \, \frac{1}{2}{\sum_{i = 1}^{M} \left\| S_i \times r_i^2 -\overline{S}\right\|_2^2},    
\label{eq:centerloss}
\end{equation}
where $M$ refers to the number of dense regions in every $T$ iterations for parameter updating and $S_i$ refers to the closeness level of region $R_i$ following Eq.~\eqref{eq:scalelevel}. 
Note that we limit the scale factor $r_i$ between 0.5 and 3 to avoid the degenerated solution of $r_i = 0$ and $\overline{S} = 0$ and potential image distortion (for too large scale factors). Meanwhile, we allow $r_i < 1$ to make the zooming-out regions also participate in the training of center loss with more samples.
It is also noteworthy to mention that we leverage the STN~\cite{jaderberg2015spatial} to achieve the scale operation.
Specifically, the scale factor of the affine matrix in the STN is set to our learned rescale factor, making the whole pipeline differentiable. Thus,
the gradients are back-propagated through $r_i$ as well as $\overline{S}$.
The derivative of $L_s$ with respect to $r_i$ is given as follows:
\begin{equation}
    \frac{\partial L_s}{\partial r_i} =2S_i (S_i \times r_i^3 - \overline{S} \times r_i).
\end{equation}

The center of closeness level $\overline{S}$ is also learnable rather than manually set. We first randomly initialize $\overline{S}$. Then, we follow the standard process of updating the center:
\begin{equation}
\Delta{\overline{S}^t} = \frac{\sum_{i=1}^{M}{(\overline{S}^t- S_i \times r_i^2})}{1+M}, \, \overline{S}^{t+T} = \overline{S}^t- \alpha \cdot \Delta{\overline{S}^t},
\label{eq:upcenter}
\end{equation}
where $\alpha$ is the learning rate for updating the center. 

The L2S is a simple yet effective module that can improve the performance of both regression-based method and localization-based method for counting in dense regions. We detail the proposed regression-based and localization-based AutoScale using L2S for crowd counting in the following.

\subsection{Regression-based counting}
\label{subsec:rbc}

\noindent\textbf{Regression model:}
\label{subsubsec:regmodel}
We first frame AutoScale in a simple manner for density map regression. Precisely, following previous works~\cite{li2018csrnet,liu2018leveraging}, we adopt a simple VGG16-based FPN~\cite{lin2017feature} as the backbone network and discard the last pooling layer and all following fully connected layers, as well as the pooling layer between \textit{stage4} and \textit{stage5} to preserve sufficient spatial information for accurate counting. As shown in Fig.~\ref{fig:pipeline}, the FPN-based backbone for counting first generates an initial density map where the estimation is relatively accurate in the sparse regions. Yet, for the dense regions, it is usually difficult to accurately regress the density maps. We propose to apply the L2S module on the dense regions to rescale all dense regions into similar closeness levels, so that the accumulated pixel values are decomposed and the distribution is transformed to be similar. To this end, we first threshold the initially predicted density map with twice of its mean density on the whole image, yielding a set of connected regions having density larger than twice of the mean value. Then we box out the maximum connected region as the selected dense region $R_d$ for each image. 

We then re-estimate the density map for the selected dense region $R_d$. For that, we crop the backbone feature on $R_d$ and pool the cropped feature to the size of $14 \times 14$, which is then fed into the L2S module. The L2S module generates a scale factor $r_d$ to rescale the dense region $R_d$. We re-estimate the density map for $R_d$ by applying the same counting network sharing parameters with the initial prediction on the rescaled region $I'_d$.

The final count is made of initially predicted density map and the re-predicted density map. Specifically, as shown in Fig.~\ref{fig:pipeline}, the final count $N_f$ is given by the sum of two counts: $N_f$ = $N_s$ +$sum(Y_d')$, where $N_s$ is the count on the sparse region (image domain excluding the selected dense region) and $sum(Y_d')$ is the refined count on the rescaled dense region $I_d'$. $N_s$ is given by the original count on the whole image $sum(Y_i)$ minus the original count on the selected dense region $sum(Y_d)$. For the regression model, the count $sum(Y)$ is obtained by integrating over the estimated density map. Note that if the size of the maximal region is smaller than a proportion $J_r$ of the input image size, no dense region is selected, implying that the underlying image mainly contains sparse regions for which the initial density map prediction is accurate enough.

\medskip
\noindent\textbf{Training objective for regression model:}
In the training phase of regression-based counting model, we follow previous density map regression methods to train the counting network. Specifically, we adopt Mean Square Error (MSE) loss function to optimize the counting network, given by
\begin{equation}
    L_m = || D - \hat{D} ||_2,
    \label{eq:mseloss}
\end{equation}
where $D$ and $\hat{D}$ are the ground-truth and predicted density map, respectively. It is noteworthy to mention that we adopt an online ground-truth update for the density map re-prediction on selected dense regions. 
Precisely, we first rescale the binary annotation in terms of points on selected dense region by multiplying the original annotated dot coordinates with the learned scale factor. 
Then we regenerate its corresponding ground-truth density map on the rescaled binary map using the same Gaussian kernel as for the initial ground-truth density map on the whole image. 

The total training objective $L_D$ for optimizing the whole density regression-based model is given by
\begin{equation}
    L_{D} = L_m^i + L_m^d + {\lambda_1} \times L_s,
    \label{eq:regloss}
\end{equation}
where $L_m^i$ and $L_m^d$ stands for the MSE loss for the initial prediction and re-prediction on the selected dense region using Eq.~\eqref{eq:mseloss}, $L_s$ is the center loss (see Eq.~\eqref{eq:centerloss}) involved in optimizing the L2S module, and $\lambda_1$ is a hyper-parameter.

\begin{figure*}[!t]
\centering
\resizebox{0.9\textwidth}{!}{
\includegraphics{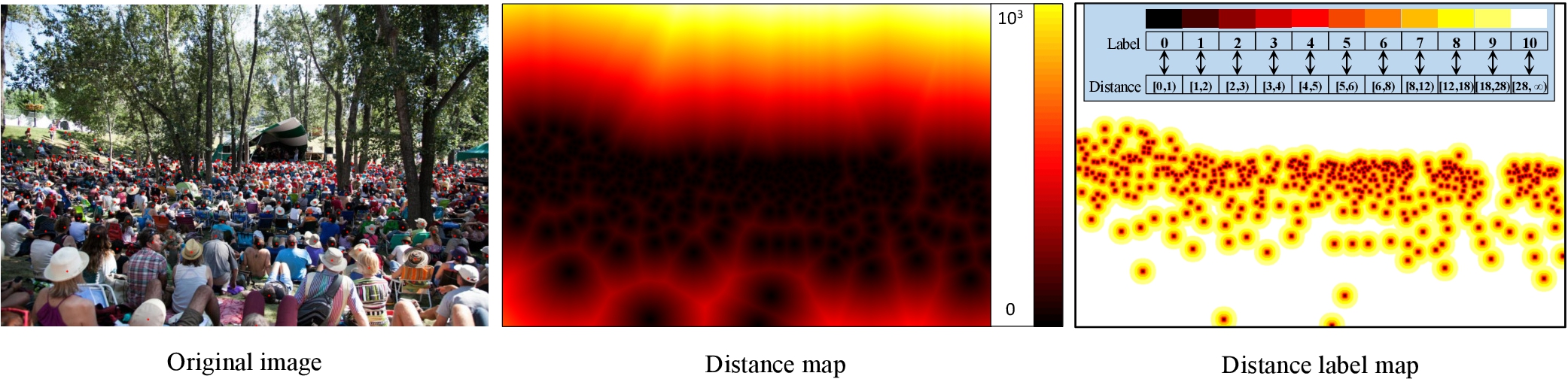}
\centering
}
\caption{Ground-truth distance-label map generation based on distance transformation from original binary location annotation $B$. The distance-label map is obtained by dividing different distances of distance map into different classes.}
\label{fig:distancemapgt}
\end{figure*}

\subsection{Localization-based counting}
\label{subsec:lbc}
Though regression-based counting provides accurate count, it does not indicate the exact person locations. Whereas, the location information is also important in many applications such as person tracking and general crowd analysis. In this section, we detail the proposed localization-based counting using a distance-label map (learned with a novel dynamic cross-entropy loss) to represent person head locations. More specifically, we first transform the binary head location annotation into a distance map through distance transformation. Then we divide the range of distances into different categories, resulting in a distance-label map (see Fig.~\ref{fig:distancemapgt}). The head locations correspond to local minima of such distance-label map. In consequence, we frame the localization-based counting problem as a dense pixel-wise classification problem, which is similar to semantic segmentation. We also resort to L2S to improve the localization accuracy in dense regions. 

\medskip
\noindent\textbf{Generation of distance-label map:}
We first apply the distance transformation~\cite{baxes1994digital} on the original annotation in terms of binary head location map. Then we classify the obtained distance map into a distance-label map $C$ by assigning different distance ranges to different classes. In this paper, the total number of distance classes $N_c$ is set to 11.
An example of generating such distance-label map $C$ is shown in Fig.~\ref{fig:distancemapgt}. The label in each pixel defines a distance level respect to its nearest head locations. Note that pixels near head locations have smaller distance label and pixels far away from head locations have larger distance label, ensuring that the distance-label blobs represent different heads without overlaps. 


Different from the existing method~\cite{idrees2018composition} that localizes Gaussian blob local maxima where the Gaussian blobs are blur and the localization is interfered by the severe overlaps between nearby heads, the adopted distance-label maps are more discriminative and there is no overlap between nearby heads. On the other hand, compared to
the binary classification method~\cite{liu2019recurrent} that directly localizes the head position, distance-label maps also roughly provides information on the closeness level via the geometrical meaning of each class, which indicates the approximate distance to the corresponding nearest head. It is noteworthy that Greg \textit{et al.}~\cite{olmschenk2019improving} also leverages the principle of distance transformation by generating iKNN maps for crowd counting. Specifically, iKNN makes use of the distance transformation for regressing a counting value. We are the first to perform person localization via distance-label map, which quantizes the continuous distance map into the discrete distance-label map.

\begin{figure*}[t!]
\centering
\includegraphics[width=0.87\textwidth]{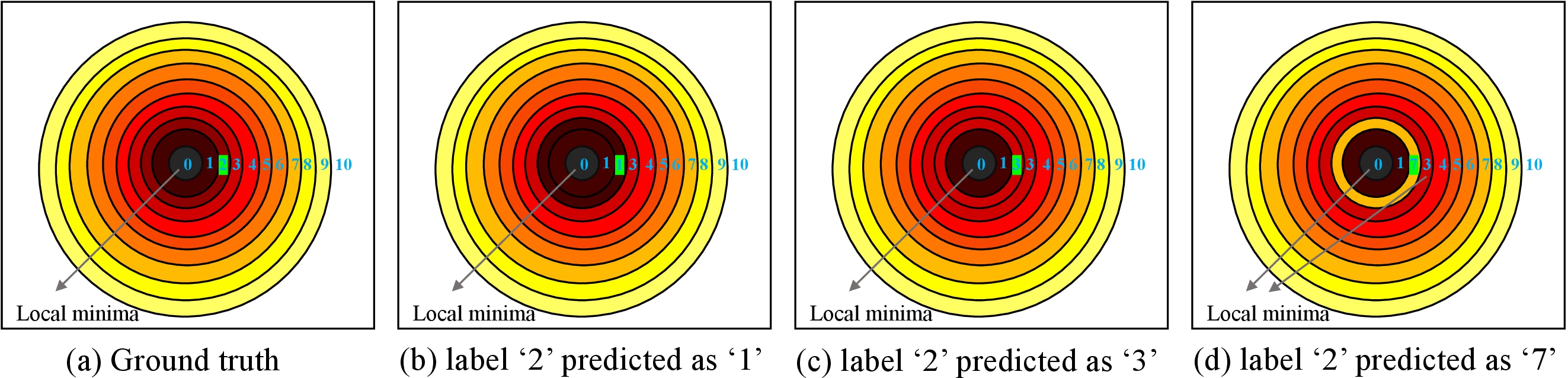}
\caption{Qualitative illustration of the purpose for DCE loss. (a) represents a local blob of distance-label map with sequential labels. (b), (c), (d) are different predicted examples with the same predicted probability on each ground-truth label.The predicted result in (b) and (c) have smaller DCE loss than (d), and preserve better the geometry structure of (a).}
\label{fig:failed_prediction}
\end{figure*}

\noindent\textbf{Localization model:}
We adopt the similar pipeline (see Fig.~\ref{fig:pipeline}) described in Section~\ref{subsec:rbc} for regression-based counting. We simply change the output target from density maps to distance-label maps. The counting network is responsible for classifying each pixel into different class labels, which provides an initial distance-label map. Similarly, this initial distance-label map prediction is accurate in sparse regions, but has difficulty in dense regions. In fact, since nearby heads in dense regions lie very close to each other, making the predicted labels prone to be the same, which hinders the accurate localization via local minima of distance-label maps. To address this, we also leverage the proposed L2S. Specifically, we threshold the initially predicted distance-label map by selecting pixels with class labels smaller than $c$ (set to 8), forming a set of candidate dense regions. We select the maximal connected region and regard its bounding box as the dense region $R_d$ for the underlying image. Similarly to the above regression-based counting with L2S, we crop the backbone feature on $R_d$ and resize it to the spatial size of $14 \times 14$. The L2S takes the resized feature and outputs a scale factor $r_d$ for rescaling $R_d$. The rescaled region image $I'_d$ is then fed into the same counting network sharing parameters with initial distance-label map prediction, leading to a re-predicted distance-label map for the selected dense region $R_d$. 

The final output is given by the sum of number of local minima in initially predicted distance-label map on sparse regions and the number of local minima in re-predicted distance-label map on the selected dense region. 
Note that we also discard the selected dense region whose size is smaller than a proportion $J_l$ of the input image size.

\begin{table*}[t]
\footnotesize
\centering
\small

\begin{tabular}{ |l|c|c| }
 \hline
 Hyper-parameter & Regression-based AutoScale & Localization-based AutoScale  \\
 \hline
 \hline
Area ratio threshold ($J_r$ and $J_l$) for dense region selection & 0.1 & 0.02 \\
Weight ($\lambda_1$ in Eq.~\eqref{eq:regloss} and $\lambda_2$ in Eq.~\eqref{eq:locloss}) of $L_s$  & 1 & 1\\
Number of iterations $T$ for updating the L2S module & 1 epoch &  1 epoch \\
Initial learning rate $\gamma$ for the whole network & 1e-7 &  1e-4 \\
Learning rate $\alpha$ for updating the center & $10^{-3}$& $10^{-3}$\\

 \hline
\end{tabular}
\caption{Settings of all involved hyper-parameters for the regression-based and localization-based AutoScale with L2S.}

\label{tab:hyper}
\end{table*}

\noindent\textbf{Training objective for localization model:}
In the training phase of localization-based counting model, we propose a dynamic cross-entropy loss function to optimize localization-based counting network. Specifically, the counting network outputs a $N_c$ channel probability map $Pr$ that classifies each pixel into a corresponding label category. We weight the cross-entropy loss based on the probability of each label category and the corresponding absolute difference with the ground-truth label value. The dynamic cross-entropy loss $L_{ce}$ for distance-label map classification is given by
\begin{equation}
\begin{tiny}
    L_{ce} \!\, =\! \,\! -\sum_{\!p \!\in \Omega\!}(\!\sum_{i \!= 0}^{\!N_c-1}(|C(p)-i|\!+1) \!\times Pr(p)^i) \times log(Pr(p)^{C(p)}),  
    \label{eq:weightceloss}
    \end{tiny}
\end{equation}

where $Pr(p)^i$ denotes the the predicted probability of pixel $p$ belonging to the $i$-th class, and $P_r(p)^{C(p)}$ is the predicted probability of pixel $p$ being the ground-truth class $C(p)$. The dynamic cross-entropy loss in Eq.~\eqref{eq:weightceloss} explicitly makes use of the distance between the prediction and the ground-truth. Indeed, each distance-label category has explicit geometrical meaning, implying the relative distance to the annotated dots. The corresponding absolute difference between each predicted label class and the ground-truth label value also measures how far is the prediction to the ground-truth. 
The relative error between the predicted and GT labels roughly reflects the relative difference between the predicted and GT distance to dots. The weighting mechanism in the DCE loss penalizes large relative difference, preserving better the geometry structure of GT distance-label map. 
For example, as shown in Fig~\ref{fig:failed_prediction}, a local blob is represented as sequential ordinary labels, \ie, 0, 1, 2, 3 ... (starting from the center). If the pixel of `2' is predicted as `1` or `3`, the localization of local minima is not influenced. Nevertheless, if the pixel of `2` is predicted as `7', it will introduce a false local minimum. Therefore, we propose to penalize more for larger difference, helping to preserve the geometry structure of the distance-label map, and providing thus more accurate localization result. It is noteworthy that for the re-prediction on selected dense regions, the ground-truth distance-label maps are regenerated online for the rescaled dense regions, in the similar way as regenerating rescaled ground-truth density maps. This ensures that the close and similar distance labels on selected dense regions can be distinguished after rescaling. 

The total training objective $L_C$ for optimizing the whole localization-based model based on distance class label map representation is given by

\begin{equation}
L_{C} = L_{ce}^i + L_{ce}^d + {\lambda_2} \times L_s, 
\label{eq:locloss}
\end{equation}
where $L_{ce}^i$ and $L_{ce}^d$ refers to the dynamic cross-entropy loss for the initial distance-label map prediction and re-prediction on the selected dense region using Eq.~\eqref{eq:weightceloss}, $L_s$ is the center loss for scale factor learning given by Eq.~\eqref{eq:centerloss} optimizing the L2S module, and $\lambda_2$ is a hyper-parameter.

\section{Experiments}
\label{sec:experiment}
\subsection{Datasets and evaluation protocol}
\label{subsec:data_eval}
We conduct experiments on the JHU-Crowd++~\cite{sindagi2020jhu}, the NWPU-Crowd~\cite{wang2020nwpu}, the UCF-QNRF~\cite{idrees2018composition} as well as the ShanghaiTech~\cite{zhang2016single} Part A and Part B datasets to demonstrate the effectiveness of both the proposed regression-based and localization-based crowd counting with L2S. Besides, we also conduct an experiment on the TRANCOS~\cite{guerrero2015extremely} dataset using the proposed distance-label map representation with dynamic cross-entropy loss, demonstrating its superiority in vehicle localization and counting.  

\begin{itemize}

\item \textbf{NWPU-Crowd~\cite{wang2020nwpu}} is currently the largest existing congest dataset with 2,133,238 annotations, containing 3109 training images, 500 val images and 1500 test images. We present the result by the provided online evaluation benchmark website.

\item \textbf{JHU-Crowd++~\cite{sindagi2020jhu}} is an extension of JHU-Crowd~\cite{sindagi2019pushing} containing 2722 training images, 500 validation images, and 1600 test images, which is collected from diverse scenarios and weather conditions. Besides, the dataset provides rich annotations, including image-level, head-level and point-level annotations. The total number of people in each image ranges from 0 to 25791. 

\item \textbf{UCF-QNRF~\cite{idrees2018composition}} is a challenging and dense dataset, containing 1201 training and 334 test high-resolution (up to $9000 \times 6000$) images. The scales of the people in this dataset vary significantly. The total number of people in each image ranges from 49 to 12865. 

\item \textbf{ShanghaiTech~\cite{zhang2016single}} consists of Part A and Part B with a total number of 1198 images. Images in Part A are scrawled from the internet, and are of different scenes and significantly varied densities. Part A is split into 300 training images and 182 test images. Part B is taken from the metropolis in Shanghai city, containing 400 images for training and 316 images for testing.

\item \textbf{TRANCOS~\cite{guerrero2015extremely}} is a vehicle counting benchmark dataset containing 1244 low resolution images captured by the publicly available video surveillance cameras in Spain. The dataset provides the split of training, validation, and test. 
  
\end{itemize}

\medskip
\textbf{Counting evaluation metrics.} We follow standard metrics widely adopted in previous works to evaluate the proposed AutoScale, including mean average error (MAE) and root mean squared error (MSE) which are defined as
\begin{equation}
MAE=\frac{1}{N_I}\sum_{i=1}^{N_I}|P_{i}-\hat{P_{i}} |, \\
MSE=\sqrt{\frac{1}{N_I}\sum_{i=1}^{N_I}|P_{i}-\hat{P_{i}}|^{2}},
\label{eq:metrics}
\end{equation}
where $N_I$ denotes the number of total images in a dataset, $P_{i}$ and $\hat{P}_i$ are the ground-truth and predicted number of people in each image.

  \begin{table*}[t]
\centering
\setlength{\tabcolsep}{0.15mm}
\resizebox{0.97\textwidth}{!}{
	\begin{tabular}{|l|c|c|cc|cc|cc|cc|cc|cc|}
		\hline
		Set&\multirow{3}{*}{Year}&\multirow{3}{*}{Backbone}& \multicolumn{2}{c|}{Val set} & \multicolumn{6}{c|}{Test set}\\
		\cline{0-0} \cline{4-11}
		Category&&&\multicolumn{2}{c|}{Overall}&\multicolumn{2}{c|}{Scene Level (only MAE)} & \multicolumn{2}{c|}{Luminance (only MAE)}&\multicolumn{2}{c|}{Overall}\\ 
		\cline{0-0} \cline{4-11}
		Method &&& MAE & MSE & Avg. & S0/S1/S2/S3/S4& Avg.&L0/L1/L2 & MAE & MSE  \\ \hline\hline
		MCNN~\cite{zhang2016single}&CVPR16&-&218.53 &700.61& 1171.9 & 356.0/72.1/103.5/509.5/4818.2 & 220.9 & 472.9/230.1/181.6 &232.5 & 714.6 \\
		SANet~\cite{cao2018scale} &ECCV18&-&171.16 &471.51& 716.3 & 432.0/65.0/104.2/385.1/\textcolor{blue}{2595.4} & 153.8 & 254.2/192.3/169.7&190.6 & 491.4\\
		PCC-Net-light~\cite{yan2019perspective} &CVPR19&-&141.37 &630.72& 944.9 & 85.3/25.6/80.4/424.2/4108.9 & 141.2 & 253.1/167.9/144.9 & 167.4 & 566.2 \\
		VGG+GPR~\cite{gao2019domain}&Tech19&VGG16&105.80 &504.40  & - & -/-/-/-/- & - &-/-/-& 127.3 & 439.9\\
		C3F-VGG~\cite{gao2019c}&Tech19&VGG16&105.79 &504.39  & 666.9 & 140.9/26.5/58.0/307.1/2801.8 & 127.9 & 296.1/125.3/91.3& 127.0 & 439.6\\
		CSRNet~\cite{li2018csrnet}&CVPR18&VGG16&104.89 &433.48& {\color{red}522.7} & 176.0/35.8/59.8/285.8/\textcolor{red}{2055.8} & 112.0 & \textcolor{green}{232.4}/121.0/95.5 &121.3 & {\color{blue}387.8}\\
		PCC-Net-VGG~\cite{yan2019perspective} &CVP19&VGG16&100.77 &573.19& 777.6 & 103.9/13.7/42.0/259.5/3469.1 & 111.0 & 251.3/111.0/82.6 & 112.3 & 457.0\\
		CAN~\cite{liu2019context} &CVPR19&VGG16&\textcolor{green}{93.58} &489.90&{\color{green}612.2} & 82.6/14.7/46.6/269.7/\textcolor{green}{2647.0} &{\color{blue}102.1} & \textcolor{blue}{222.1}/104.9/82.3& 106.3 & {\color{red}386.5}\\
		\hline
		SFCN\dag~\cite{wang2019learning}&CVPR19&ResNet101&95.46 &608.32& 712.7 & \textcolor{red}{54.2}/14.8/44.4/\textcolor{green}{249.6}/3200.5 & {\color{green}106.8} & 245.9/\textcolor{green}{103.4}/\textcolor{green}{78.8} & {\color{green}105.7} & 424.1\\
		BL~\cite{ma2019bayesian}&ICCV19&VGG19&93.64 &\textcolor{green}{470.38}& 750.5 & \textcolor{blue}{66.5}/\textcolor{red}{8.7}/\textcolor{green}{41.2}/249.9/3386.4 & 115.8 & 293.4/\textcolor{blue}{102.7}/\textcolor{blue}{68.0}
		 &{\color{blue}105.4}  &454.2\\
		\hline
		FPN (\textbf{ours}) &-&VGG16&\textcolor{blue}{70.3} &\textcolor{blue}{364.5}&691.9 &163.2/\textcolor{green}{13.8}/\textcolor{blue}{39.8}/\textcolor{blue}{247.7}/2995.0 & 111.5 &261.2/105.3/80.4 &108.3 &469.1\\
		AutoScale (\textbf{ours}) &-&VGG16&{\color{red}68.8}&{\color{red}356.9}& \textcolor{blue}{608.2} & \textcolor{green}{81.4}/\textcolor{blue}{11.3}/\textcolor{red}{38.1}/\textcolor{red}{226.0}/2683.7 & \textcolor{red}{94.3}& \textcolor{red}{220.9}/\textcolor{red}{93.4}/\textcolor{red}{65.5}&\textcolor{red}{94.1}&\textcolor{green}{388.2}\\
		\hline
		\hline
	   RAZ\_localization+*~\cite{liu2019recurrent} &CVPR19&VGG16&128.7&665.4&1166.0&60.6/17.1/48.3/364.7/5339.0 & 153.1&350.7/147.5/115.2 &151.4& 634.6\\
		\hline
        FPN* (\textbf{ours}) &-&VGG16&111.6&605.3&1074.0&43.8/\textcolor{red}{13.4}/47.0/360.2/4905.5 & 147.3&345.4/139.7/105.0 &143.2& 603.2\\
		AutoScale* (\textbf{ours}) &-&VGG16&{\color{red}97.3}&{\color{red}571.2}& \color{red}{871.2}&\textcolor{red}{42.3}/18.8/\textcolor{red}{46.1}/\textcolor{red}{301.7}/\textcolor{red}{3947.0} & \color{red}{127.1} & \textcolor{red}{301.3}/\textcolor{red}{122.2}/\textcolor{red}{86.0}&\color{red}{123.9}&\color{red}{515.5}\\
		\hline
	\end{tabular}}
	
\caption{Quantitative results on val/test set of NWPU-Crowd dataset. Scene level is divided into five different ranges:0, (0,100], (100,500], (500,5000], and \textgreater 5000. L0, L1 and L2 denote three luminary levels respectively:[0,0.25], (0.25,5], (0.5,0.75]. Red, blue, and green color respectively indicate the first, second and third place in this table. The methods with $*$ denotes the localization-based methods.}
\label{tab:NWPU_result}
\end{table*}
\textbf{Localization evaluation metrics.}
We calculate the Precision, Recall, and F-measure metrics to evaluate the localization performance. Specifically, when the distance between a given predicted point $p_p$ and ground truth point $p_g$ is less than a distance threshold $\sigma$, it means that $p_p$ and $p_g$ are successfully matched. For ShanghaiTech Part A, we calculate the localization metrics with four different $\sigma$ (4, 8, 16, and KNN distance). For the UCF-QNRF dataset, we report the average precision, average recall, and average F-measure at different distance tolerance thresholds $\sigma$: $(1, 2, 3, \dots, 100)$ pixels, following~\cite{idrees2018composition}. For the NWPU-Crowd dataset~\cite{wang2020nwpu}, we follow~\cite{wang2020nwpu} that chooses two adaptive thresholds $\sigma_s = min(w,h)/2$ and $\sigma_l={\sqrt {{w^2} + {h^2}}}/2 $ for each individual person, where $w$ and $h$ is the width and height of the annotated bounding box for the corresponding person. 
Note that we use the implementation
code\footnote{https://github.com/gjy3035/NWPU-Crowd-Sample-Code-for-Localization} provided by~\cite{wang2020nwpu} to compute the localization evaluation metrics for all datasets.

\begin{table*}[t]
\normalfont
\centering
\setlength{\tabcolsep}{0.10mm}
\resizebox{0.97\textwidth}{!}{
\begin{tabular}{|l|c|c|cc|cc|cc|cc|cc|cc|cc|cc|}
	\hline
	Set&\multirow{3}{*}{Year} &\multirow{3}{*}{Backbone} &\multicolumn{8}{c|}{Val set} & \multicolumn{8}{c|}{Test set}\\
	\cline{0-0} 	\cline{4-19} 
	Category &&& \multicolumn{2}{c|}{Low} & \multicolumn{2}{c|}{Medium}& \multicolumn{2}{c|}{High}&\multicolumn{2}{c|}{Overall}&\multicolumn{2}{c|}{Low} & \multicolumn{2}{c|}{Medium}& \multicolumn{2}{c|}{High}&\multicolumn{2}{c|}{Overall}\\
	\cline{0-0} 	\cline{4-19} 
	Method &&& MAE & MSE & MAE & MSE& MAE& MSE& MAE & MSE& MAE & MSE & MAE & MSE& MAE& MSE& MAE & MSE \\ \hline\hline
	MCNN~\cite{zhang2016single} &CVPR16&-& 90.6& 202.9& 125.3& 259.5& 494.9 & 856.0 & 160.6& 377.7& 97.1 & 192.3& 121.4& 191.3& 618.6& 1,166.7 & 188.9& 483.4 \\  
	CMTL~\cite{sindagi2017cnn} &AVSS17&-& 50.2& 129.2& 88.1 & 170.7& 583.1 & 986.5& 138.1& 379.5 & 58.5& 136.4& 81.7 & 144.7 & 635.3& 1,225.3 & 157.8 &490.4\\  
	DSSI-Net~\cite{liu2019crowd}&ICCV19&VGG16& 50.3& 85.9 & 82.4 & 164.5& 436.6 & 814.0 & 116.6& 317.4 & 53.6& 112.8  & 70.3& 108.6  & 525.5& 1,047.4 & 133.5  & 416.5\\  
	CAN~\cite{liu2019context}&CVPR19&VGG16& 34.2& 69.5 & 65.6 & 115.3& 336.4 & 619.7 & 89.5 & 239.3& 37.6& 78.8& 56.4& 86.2& 384.2& 789.0& 100.1& 314.0\\  
	SANet~\cite{cao2018scale} &ECCV18&-& 13.6& 26.8 & 50.4 & 78.0 & 397.8 & 749.2& 82.1 & 272.6 & 17.3& 37.9& 46.8& 69.1& 397.9& 817.7& 91.1& 320.4\\ 
	CSR-Net~\cite{li2018csrnet} &CVPR18&VGG16& 22.2& 40.0 & 49.0 & 99.5 & 302.5 & 669.5 & 72.2 & 249.9 & 27.1& 64.9& 43.9& 71.2& 356.2& 784.4& 85.9& 309.2\\  
	CG-DRCN~\cite{sindagi2020jhu}&PAMI20 &VGG16& 17.1& 44.7 & 40.8 & 71.2      & 317.4 & 719.8 & 67.9 & 262.1 & 19.5& 58.7& 38.4& 62.7& 367.3& 837.5&82.3& 328.0\\
	MBTTBF~\cite{sindagi2019multi} &ICCV19 &VGG16& 23.3& 48.5 & 53.2 & 119.9& 294.5 & 674.5 & 73.8 & 256.8 & 19.2& 58.8& 41.6& 66.0& 352.2& 760.4& 81.8&299.1\\  
	SFCN~\cite{wang2019learning} &CVPR19&VGG16& 11.8& 19.8 & {39.3}& 73.4 & 297.3 & 679.4&62.9 & 247.5 & 16.5& 55.7& 38.1& 59.8& 341.8 &758.8 & 77.5&297.6\\  
	\hline
	BL~\cite{ma2019bayesian}& ICCV19&VGG19& \textcolor{red}{6.9}&\textcolor{red}{10.3}& 39.7 & 85.2 &279.8& 620.4&  {\color{green}59.3 }& 229.2 &\textcolor{red}{10.1}&\textcolor{red}{32.7}& \textcolor{green}{34.2}&\textcolor{green}{54.5}& 352.0& 768.7&75.0& 299.9\\  
	CG-DRCN~\cite{sindagi2020jhu}&PAMI&ResNet101& 11.7&\textcolor{green}{24.8} &\textcolor{blue}{35.2} & \textcolor{red}{57.5} & {\color{green}273.9} & 676.8  & {\color{red}57.6}& 244.4& \textcolor{blue}{14.0}&\textcolor{blue}{42.8}& 35.0& 53.7& 314.7& {\color{green}712.7}& {\color{green}71.0}& {\color{green}278.6} \\ \hline
	FPN (\textbf{our})&-&VGG16&11.9&\textcolor{blue}{22.3}&46.7&75.8&303.5&629.2&67.5&230.5&15.9&54.1&45.4&94.1&342.4&762.4&81.5&303.9\\
	AutoScale (\textbf{ours}) &-&VGG16&15.9&28.8&44.0&86.1&{\color{blue}272.6}&{\color{blue}580.0}&63.4&{\color{blue}216.0}&22.2&70.6&42.5&93.0&307.4&729.5&76.4&292.7 \\
	\hline
	FPN (\textbf{our})&-&VGG19&\textcolor{green}{11.1}&26.0&\textcolor{green}{37.2}&\textcolor{blue}{69.5}&291.0&\textcolor{green}{591.2}&60.5& \color{green}{224.8} &16.0&\textcolor{green}{44.4}&\textcolor{blue}{33.4}&\textcolor{red}{51.5}&\textcolor{blue}{298.6}&\textcolor{blue}{684.3}&\textcolor{blue}{68.1}& \textcolor{blue}{267.8}\\
	AutoScale (\textbf{ours}) &-&VGG19&\textcolor{blue}{10.8}&25.2&\textcolor{red}{34.9}&\textcolor{green}{69.9}&\textcolor{red}{270.2}&\textcolor{red}{567.2}&\color{blue}{58.2}& \textcolor{red}{212.4} &\textcolor{green}{14.6}&46.7&\textcolor{red}{33.0}&\textcolor{blue}{52.8}&{\color{red}287.2}&{\color{red}675.6}&{\color{red}65.9}&{\color{red}264.8} \\
	
	\hline\hline
	LSC-CNN*~\cite{sam2020locate} &TPAMI20&VGG16& 6.8 &10.1& 39.2&64.1& 504.7 & 860.0 & 87.3 & 309.0 &10.6 & 31.8 &34.9 & 55.6& 601.9& 1,172.2 & 112.7 & 454.4\\ 
	\hline
	FPN* (\textbf{ours})&-&VGG16&10.2&\textcolor{red}{14.8} &35.4&55.8 &459.7&894.5 &80.4&320.2&\textcolor{red}{13.0}&\textcolor{red}{28.4}&33.1&56.3&526.0&1,090.4&100.9&423.0\\
	AutoScale* (\textbf{ours}) &-&VGG16&\textcolor{red}{10.0}&15.3 &\textcolor{red}{33.5}&\textcolor{red}{54.2}&{\color{red}351.7}&{\color{red}720.3}& {\color{red}65.7}& {\color{red}258.9}&13.2&30.2&\textcolor{red}{32.3}&\textcolor{red}{52.8}&{\color{red}425.6}&{\color{red}916.5}& {\color{red}85.6}& {\color{red}356.1} \\
	\hline
\end{tabular}}
\caption{Quantitative results on the val/test sets of JHU-Crowd++ dataset. "Low", "Medium" and "High" respectively indicates three categories based on different ranges:[0,50], (50,500], and \textgreater 500. Red, blue, and green color respectively indicates the first, second and third place in this table. The methods with $*$ denote the localization-based methods.}
\label{tab:JHU_result}
\end{table*}

\begin{figure*}
\centering
\resizebox{0.9\textwidth}{!}{
    \includegraphics{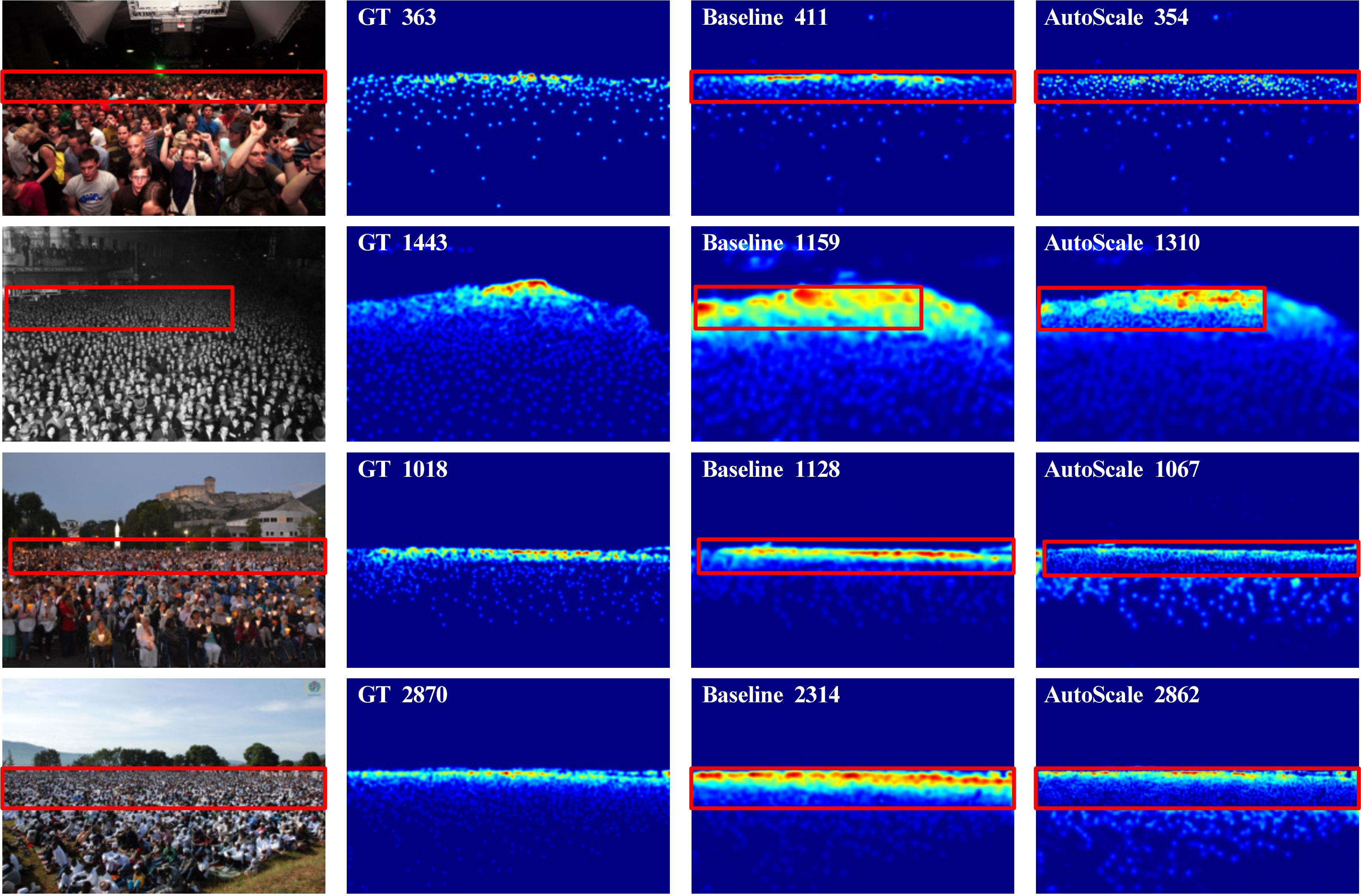}
    }
    \caption{Qualitative visualization of density maps generated by the proposed AutoScale. From left to right: original images, ground-truth density maps, baseline results, and results with L2S. The enclosed regions are the automatically selected dense regions, which are rescaled via L2S for re-prediction.}
\centering
\label{fig:regvis}
\end{figure*}

\subsection{Implementation details}
\label{subsec:implementation}
For NPWU-CROWD, JHU-Crowd++, and UCF-QNRF dataset, we resize all the images by setting the longer side size to 2048 while maintaining the corresponding ratio. The original image size is used on the ShanghaiTech Part A and Part B dataset.
We follow the procedure described in Section~\ref{subsec:adm} to generate ground-truth density maps for the regression-based AutoScale with different spread parameters, which are set to 4, 8, 8, 6, and 8 on the ShanghaiTech Part A, Part B, UCF-QNRF, NWPU-Crowd, JHU-Crowd++ datasets, respectively. For the localization-based AutoScale, we generate the ground-truth distance-label maps as described in Section~\ref{subsec:lbc} and depicted in Fig.~\ref{fig:distancemapgt} for all involved datasets. 

During the training phase, original or downsampled images are fed to the network with data augmentations, including horizontal flipping, adding noise, and random scaling. The settings for all involved hyper-parameters are depicted in Tab.~\ref{tab:hyper}. Specifically, the area ratio threshold $J$ for discarding small dense regions is set to 0.1 and 0.02 for the regression and localization-based AutoScale, respectively. The weight of the center loss $L_s$ on scale factors $\lambda_1$ in Eq.~\eqref{eq:regloss} and $\lambda_2$ in Eq.~\eqref{eq:locloss} involved in training objective for regression-based and localization-based AutoScale are both set to 1. The parameters in L2S are initialized with Gaussian random values and are updated each epoch.

We use Adam~\cite{kingma2014adam} to optimize both training objectives with batch size set to 1. The weight decay is set to $5\times 10^{-4}$. The learning rate $\alpha$ for updating the center in Eq.~\eqref{eq:upcenter} is set to $10^{-3}$. The learning rate $\gamma$ for the whole network is set to $10^{-7}$ and $10^{-4}$ for the regression-based and localization-based AutoScale, respectively.
During the test phase, we use the same area ratio threshold $J$ as the training phase. The proposed AutoScale is implemented in Pytorch. All experiments are carried out on a workstation with an Intel Xeon 16-core CPU (3.5GHz), 32GB RAM, and a single Tesla V100 GPU. 
\begin{table*}[t]
\centering
\begin{tabular}{ |l|c|c|cc|cc|cc|cc| }
 \hline
 {\multirow{2}{*}{Method}} &{\multirow{2}{*}{Year}}&{\multirow{2}{*}{Backbone}}&\multicolumn{2}{c|}{UCF-QNRF}  & \multicolumn{2}{c|}{ShanghaiTech Part A} & \multicolumn{2}{c|}{ShanghaiTech Part B} \\
\cline{4-9}
&&& MAE & MSE & MAE & MSE&MAE&MSE \\
 \hline\hline
     MCNN~\cite{zhang2016single} &CVPR16&-&277.0&426.0& 110.2 & 173.2 & 26.4 & 41.3  \\
     CMTL~\cite{sindagi2017cnn} &AVSS17&-& 252.0&514.0& 101.3 &152.4 &20.0&31.1\\
     SANet~\cite{cao2018scale} &ECCV18 &-&-&-& 67.0 &104.5  & 8.4 & 13.6\\
     CL~\cite{idrees2018composition}&ECCV18&-&132.0&191.0&-&-&-&-\\
     CSRNet~\cite{li2018csrnet} &CVPR18&VGG16&-&-& 68.2 & 115.0  & 10.6&16.0 \\
     SL2R~\cite{liu2019exploiting} &TPAMI19&VGG16&124.0&196.0& 73.6 & 112.0 & 13.7 & 21.4  \\
     PACNN+CSRNet~\cite{shi2019revisiting}&CVPR19&VGG16&-&-&62.4&102.0&7.6& 11.8\\
     CFF~\cite{shi2019counting} &ICCV19 &-&- &-&65.2 &109.4 &	7.2 & 12.2 \\
     SPANet+SANet~\cite{cheng2019learning} &ICCV19&-&- &-&59.4&\color{blue}{92.5} &\color{blue}{6.5} &\color{red}{9.9} \\
     HA-CNN~\cite{sindagi2019ha}&TIP19&VGG16&118.1&180.4&62.9&94.9&8.1&13.4\\
     RAZ\_fusion~\cite{liu2019recurrent}&CVPR19&VGG16&116.0&195.0&65.1 & 106.7 & 8.4 & 14.1\\
     PGCNet~\cite{yan2019perspective} &ICCV19&VGG16&-&-&57.0&\color{red}{86.0}&8.8&13.7\\
     TEDnet~\cite{jiang2019crowd}&CVPR19&-&113.0&188.0&64.2&109.1&8.2&12.8\\
     CG-DRCN~\cite{sindagi2020jhu}&PAMI20&VGG16&112.2&176.3&64.0&98.4&8.5&14.4\\
     RANet~\cite{zhang2019relational}&ICCV19 &-&111.0&190.0&59.4 & 102.0 &7.9 & 12.9\\
     CAN~\cite{liu2019context}&CVPR19&VGG16&107.0&183.0&62.3&100.0&7.8&12.2\\
     S-DCNet~\cite{xiong2019open} &ICCV19&VGG16 &104.4&176.1&58.3 & 95.0 & \color{green}{6.7} & 10.7 \\
     AMSNet~\cite{hu2020count}
     &ECCV20&-&101.8&163.2&\textcolor{green}{56.7}&\textcolor{green}{93.4}&\textcolor{green}{6.7}&\textcolor{blue}{10.2}\\
     HyGnn~\cite{luo2020hybrid} &AAAI20&VGG16&100.8&185.3&60.2&94.5&7.5&12.7\\
     SDANet~\cite{miaoshallow} &AAAI20 &-&-&-&63.6&101.8&7.8&\color{blue}{10.2}\\
     RPNet~\cite{yang2020reverse} &CVPR20&VGG16&-&-&61.2&96.9&8.1&11.6\\
     DSSI-Net~\cite{liu2019crowd}&ICCV19 &VGG16&99.1&159.2&60.6&96.0 &6.8&10.3\\
     MBTTBF-SCFB~\cite{sindagi2019multi} &ICCV19&VGG16 &97.5&165.2&60.2&94.1&-&-\\
     PaDNet~\cite{tian2019padnet} &TIP19 &VGG16&96.5 & 170.2&59.2 & 98.1 & 8.1 & 12.1 \\
     ASNet~\cite{jiang2020attention} &CVPR20&VGG16&91.5&159.7&57.7&90.1&-&-\\
     LibraNet~\cite{liu2020WeighingCounts} &ECCV20&VGG16&88.1&\textcolor{blue}{143.7}&\textcolor{blue}{55.9}&97.1&7.3&11.3\\
     AMRNet~\cite{liu2020adaptive} &ECCV20&VGG16&\textcolor{blue}{86.6}&152.2&61.5&98.3&7.0&11.0\\
     ADSCNet~\cite{bai2020adaptive} &CVPR20&VGG16&\textcolor{red}{71.3}&  \textcolor{red}{132.5}&\textcolor{red}{55.4}&97.7&\textcolor{red}{6.4}&11.3\\
     \hline
    SFCN\dag~\cite{wang2019learning}&CVPR19&ResNet101&102.0&171.4&64.8&107.5&7.6&13.0\\
     CG-DRCN~\cite{sindagi2020jhu}&PAMI20&ResNet101&95.5&164.3&60.2&94.0&7.5&12.1\\
     BL~\cite{ma2019bayesian} &ICCV19 &VGG19&88.7&154.8& 62.8& 101.8& 7.7 & 12.7\\
     \hline
     FPN (\textbf{ours}) &-&VGG16&92.0&159.1& 66.1 &  111.9  &8.0   &12.0 \\
     AutoScale (\textbf{ours})&-&VGG16& \textcolor{green}{87.5} & \textcolor{green}{147.8}& 60.5&100.4 &6.8 &11.3 \\
     \hline
     \hline

    Method* in~\cite{ribera2019}&CVPR19 &VGG16& 258.6 & 499.6 & 129.7  & 189.6  & 19.9& -  \\
    LCFCN*~\cite{laradji2018blobs}&ECCV18 &VGG16 & 249.3 & 525.6 &121.6 & 223.5 & 13.1 & - \\
    PSDDN*~\cite{liu2019point}&CVPR19&ResNet101& -&-& 65.9 &112.3 &9.1&14.2\\
    LSC-CNN*~\cite{sam2020locate}&TPAMI20&VGG16&120.5&218.2& 66.4 & 117.0 &\color{red}{8.1} &\color{red}{12.7}\\
     RAZ\_localization+*~\cite{liu2019recurrent}&CVPR19 &VGG16&118.0&198.0& 71.6 & 120.1 & 9.9 & 15.6  \\

     \hline
     FPN* (\textbf{ours})  &-&VGG16&124.8&234.7& 75.7 & 150.4  & 10.4  &18.8\\
     AutoScale* (\textbf{ours})&-&VGG16&\color{red}{104.4}&\color{red}{174.2}& \color{red}{65.8}&\color{red}{112.1} &8.6& 13.9 \\
     \hline
\end{tabular}
\caption{Quantitative comparison (in terms of MAE and MSE) of the regression-based AutoScale and localization-based AutoScale with state-of-the-art methods on three widely adopted benchmark datasets. The methods with $*$ represents localization-based methods. Red, blue, and green color respectively indicate the first, second and third place in this table.}
\label{tab:maemse}
\end{table*}

\begin{figure*}
\centering
\resizebox{0.9\textwidth}{!}{
    \includegraphics{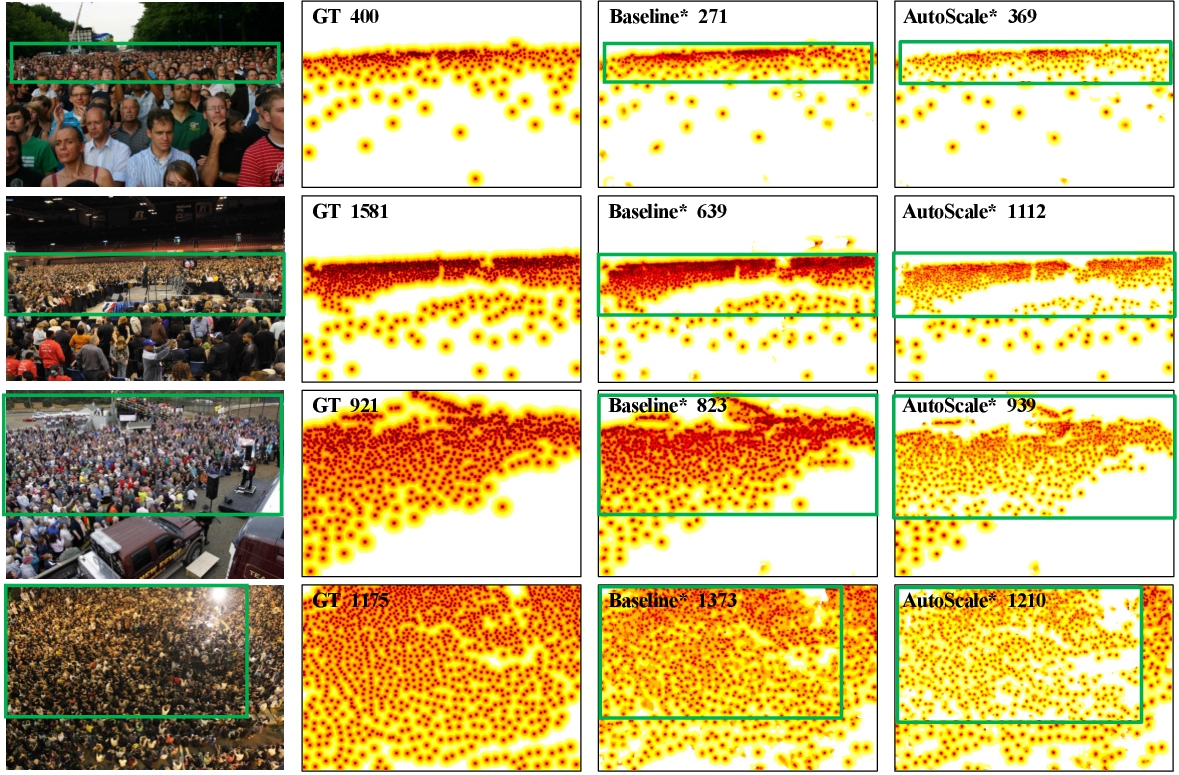}
    }
    \caption{Qualitative visualization of distance-label maps given by the proposed AutoScale. From left to right: original images, ground-truth distance-label maps, baseline results, and results with L2S. The enclosed regions are the automatically selected dense regions, which are rescaled via L2S for re-prediction.}
\centering
\label{fig:locvis}
\end{figure*}
\begin{figure*}
\centering
\resizebox{0.9\textwidth}{!}{
\includegraphics{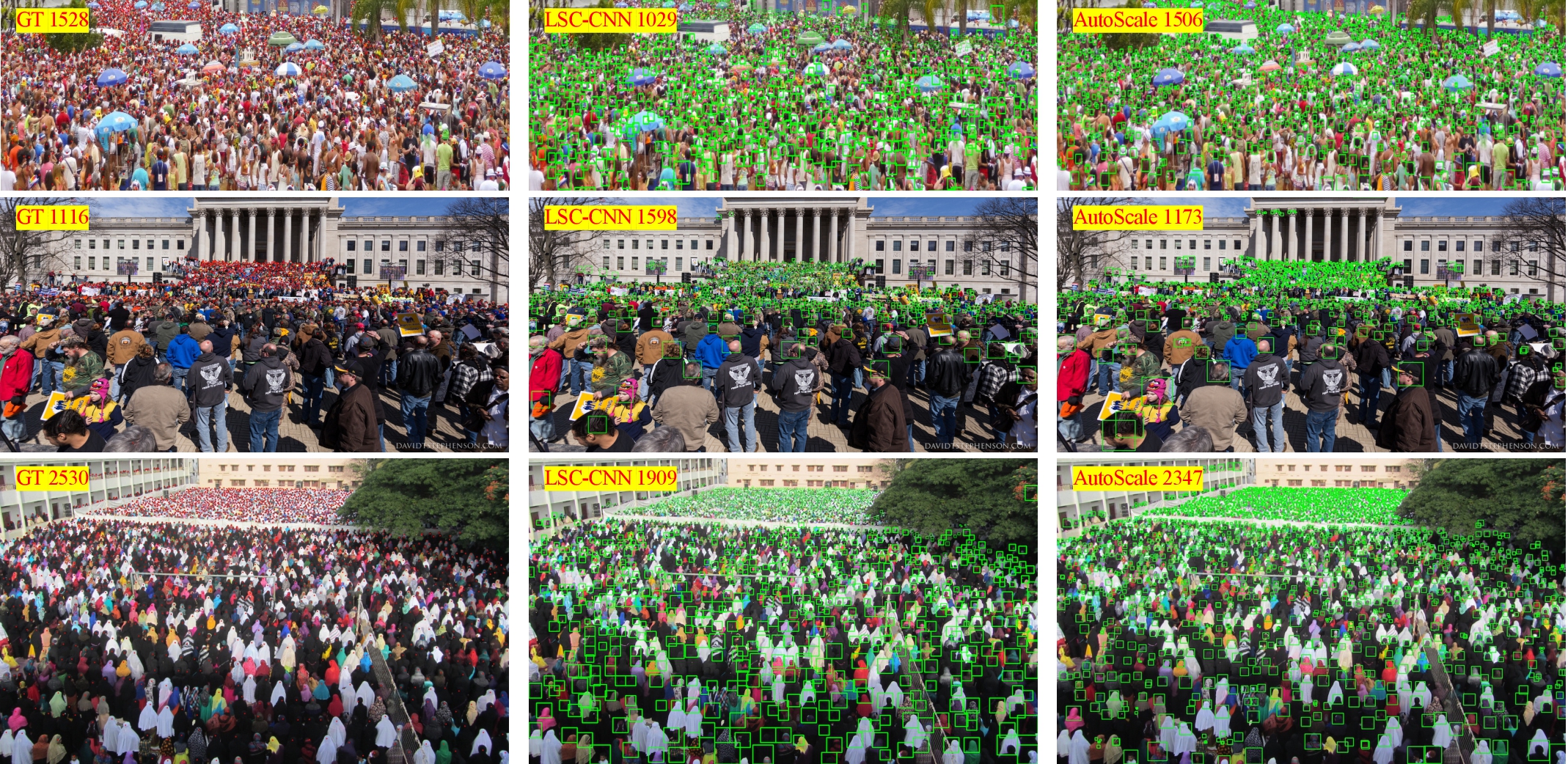}
}
\caption{Qualitative visualization of detected person locations by the localization-based AutoScale. Red points are the ground-truth. To more clearly present our localization results, we generate bounding boxes (green boxes) according to the KNN distance of each point, which follows and compares with LSC-CNN~\cite{sam2020locate}. }
\label{fig:locationvis}
\centering
\end{figure*}

\begin{table*}[t]
\footnotesize
\small
\centering
\setlength{\tabcolsep}{0.97mm}

\begin{tabular}{ |c|ccc|ccc|ccc|ccc| }
 \hline
 {\multirow{2}{*}{Method}} & \multicolumn{3}{c|}{$\sigma$ = KNN distance} & \multicolumn{3}{c|}{$\sigma$ = 4}&\multicolumn{3}{c|}{$\sigma$ = 8}&\multicolumn{3}{c|}{$\sigma$ = 16} \\
\cline{2-13}
 {} & P (\%) & R (\%) &F (\%)& P (\%)& R (\%) &F (\%) &P (\%) &R (\%)&F (\%)&P (\%) & R (\%)&F (\%) \\
\hline
{LCFCN\cite{laradji2018blobs}} &  \textbf{87.7\%}& 52.6\%&65.8\% &43.3\% &26.0\% &32.5\%& \textbf{75.1\%} &45.1\% &56.3\% &  \textbf{87.0\%}& 52.2\%&65.3\%\\
{Method in~\cite{ribera2019}} & 85.7\% & 53.0\%&65.4\% &34.9\% &20.7\% &25.9\%& 67.7\% &44.8\%& 53.9\% & 84.5\%& 51.8\%&64.2\%\\
{LSC-CNN~\cite{sam2020locate}} & 82.8\%& 79.1\%&80.9\% &33.4\% &31.9\% & 32.6\%&63.9\% &61.0\% &62.4\% & 83.9\%&80.1\% & 81.9\%\\
FPN* (\textbf{ours}) &83.9\%&78.4\% &81.1\% &55.1\% & 51.5\%& 53.3\%&74.4\% &69.6\% &71.9\% &84.5\%&79.1\%&81.7\% \\
AutoScale* (\textbf{ours})&83.5\%&\textbf{80.5\%}&\textbf{82.0\%}&\textbf{56.2\%}&\textbf{54.2\%}&\textbf{55.2\%}&74.4\%&\textbf{71.7\%} &\textbf{73.0\%}&84.4\%&\textbf{81.4\%}&\textbf{82.9\%} \\
\hline
\end{tabular}
\caption{Quantitative evaluation of localization-based methods on the ShanghaiTech Part A dataset using Precision (P), Recall (R), and F-measure (F) at different distance tolerance thresholds $\sigma$ for one-to-one correct match between the predicted and ground-truth localization result.}
\label{tab:ShanghaiA_loc}
\end{table*}

\subsection{Experimental comparison}
\label{subsec:expcomp}

\textbf{Regression-based crowd counting result:} We first evaluate the regression-based AutoScale using L2S. Fig.~\ref{fig:regvis} presents some sample qualitative density maps. Qualitatively, the proposed L2S helps to improve the density map prediction on dense regions, boosting the count accuracy. The quantitative comparison with state-of-the-art methods on NWPU-Crowd, JHU-Crowd++, ShanghaiTech, and UCF-QNRF datasets is depicted in Tab.~\ref{tab:NWPU_result}, Tab.\ref{tab:JHU_result} and Tab.~\ref{tab:maemse}, respectively.

Our VGG16-based AutoScale outperforms the other VGG16-based methods (including VGG16-based CG-DRCN \cite{sindagi2020jhu}) on JHU-Crowd++ and NWPU-Crowd dataset, and is very competitive to the other methods including most VGG16-based related CVPR/ECCV/AAAI 2020 methods on UCF-QNRF, ShanghaiTech Part A and ShanghaiTech Part B datasets. Specifically, AutoScale has superior performance on the extremely dense set, e.g., the ``High'' part of JHU-Crowd++ dataset and the ``S4'' in NWPU-Crowd dataset, which demonstrates the effectiveness of the proposed L2S to dense regions in crowd counting. On the relative sparse datasets such as ShanghaiTech Part A and Part B, L2S still improves the baseline model significantly.

It is noteworthy that
the state-of-the-art methods BL \cite{ma2019bayesian} and CG-DRCN \cite{sindagi2020jhu} leverage VGG19 and ResNet101 as the backbone, which is stronger than our adopted VGG16 backbone on all datasets. For a fair
comparison on the JHU-Crowd++ dataset, We also implement our AutoScale with VGG19
backbone on the JHU-Crowd++ dataset, and observe that the VGG19-based AutoScale achieves 65.9 MAE and 264.8 MSE on the JHU-Crowd++ dataset, improving VGG19-based BL \cite{ma2019bayesian} (resp. ResNet101-based CG-DRCN \cite{sindagi2020jhu}) by 9.1 MAE (resp. 5.1) and 35.1
(13.8) MSE. Besides, as depicted in Tab.~\ref{tab:NWPU_result} and~\ref{tab:maemse}, despite the less strong backbone network, our VGG16-based AutoScale outperforms VGG19-based BL~\cite{ma2019bayesian} and ResNet101-based CG-DRCN~\cite{sindagi2020jhu} by 1.2 (\resp 7.0) and 8.0 (\resp 16.5) MAE (\resp MSE) on the UCF-QNRF dataset, respectively, and improves VGG19-based BL~\cite{ma2019bayesian} and ResNet101-based SFCN\dag~\cite{wang2019learning} by 11.3 (\resp 66.0) and 11.6 (\resp 35.9) MAE (\resp MSE) on the NWPU-Crowd dataset, respectively. On the UCF-QNRF dataset, AutoScale performs slightly worse than AMRNet~\cite{liu2020adaptive} and worse than ADSCNet~\cite{bai2020adaptive}. Compared with AMRNet which adopts some specific designs such as multi-scale fusion, and multi-activation fusion, AutoScale is simply to automatically zoom in the dense regions for refinement without other tricks, and achieves slightly worse MAE than AMRNet~\cite{liu2020adaptive} (87.5 VS 86.6 MAE) but better MSE (147.8 VS 152.2 MSE). Compared with the ADSCNet~\cite{bai2020adaptive} which proposes the adaptive dilations and ground-truth correction mechanism (bringing 18.8 MAE improvement), AutoScale targets at a different perspective, \textit{i.e.}, a general module to refine the prediction on the dense region.


It is also noteworthy that the proposed method performs better on dense crowds than sparse ones, which further confirms that the learning to scale module works well. Since the proposed method is dedicated to improving the counting accuracy on dense regions causing long-tailed distribution problem, it is reasonable that the performance improvement on the sparse dataset is not as good as that on dense datasets. Overall, the proposed method improves the performance on very crowded scenes (such as large gatherings, train stations, stadiums), and does not harm the performance on sparse scene that does not suffer from long-tailed distribution issue.

\begin{table}[t]
\small
\footnotesize
\centering
\setlength{\tabcolsep}{1mm}
\resizebox{0.47\textwidth}{!}{
\begin{tabular}{|c|c|c|c|}
\hline
Method &Av.Precision&Av.Recall&F-measure\\
\hline
MCNN~\cite{zhang2016single}&59.93\%&63.50\%&61.66\%\\
ResNet74~\cite{he2016deep}&61.60\%&66.90\%&64.14\%\\
DenseNet63~\cite{huang2017densely}&70.91\%&58.10\%&63.87\%\\
Encoder-Decoder~\cite{badrinarayanan2017segnet}&71.80\%&62.98\%&67.10\%\\
CL~\cite{idrees2018composition}&75.80\%&59.75\%&66.82\%\\
LCFCN~\cite{laradji2018blobs}&77.89\%&52.40\%&62.65\%\\
Method in~\cite{ribera2019}&75.46\%&49.87\% & 60.05\%\\
LSC-CNN\cite{sam2020locate} &75.84\% &74.69\% & 75.26\%\\
\hline
  FPN* (\textbf{ours}) & \textbf{81.40}\%&74.03\% &77.54\%\\
  AutoScale* (\textbf{ours})& 81.31\%&\textbf{75.75}\%&\textbf{78.43}\%\\
 \hline
\end{tabular}}
\caption{Quantitative evaluation of localization-based methods on the UCF-QNRF dataset. We report the average precision, average Recall, and average F-measure at different distance thresholds $\sigma$: $(1, 2, 3, \dots, 100)$ pixels.}
\label{tab:ucf_localization_metrics}
\end{table}

\begin{table*}[t]
    \small
	\vspace{-0.25cm}
	\footnotesize
	\centering
	\begin{tabular}{|c|c|c|c|c|c|}
		\hline
		\multirow{2}{*}{Method}	&\multirow{2}{*}{Backbone} &\multicolumn{2}{c|}{Val}  &\multicolumn{2}{c|}{Test}  \\ %
		\cline{3-6}
		& & \textbf{F}/P/R (\%)  &MAE/MSE & \textbf{F}/P/R (\%) &MAE/MSE\\
		\hline
		\multirow{2}{*}{Faster RCNN~\cite{ren2015faster}}  &\multirow{2}{*}{ResNet-101}  &$\sigma_l$: 7.3/\textbf{96.4}/3.8 &\multirow{2}{*}{377.3/1051.2} &$\sigma_l$: 6.7/\textbf{95.8}/3.5 &\multirow{2}{*}{414.2/1063.7}  \\
		\cline{3-3}	\cline{5-5}
		&&$\sigma_s$: 6.8/\textbf{90.0}/3.5& &$\sigma_s$: 6.3/\textbf{89.4}/3.3&\\
		\hline
		
		\multirow{2}{*}{TinyFaces~\cite{hu2017finding}}  &\multirow{2}{*}{ResNet-101}  &$\sigma_l$: 59.8/54.3/66.6 & \multirow{2}{*}{240.4/736.2} &$\sigma_l$: 56.7/52.9/\textbf{61.1} &\multirow{2}{*}{272.4/764.9}     \\
		\cline{3-3}	\cline{5-5}
		&&$\sigma_s$: 55.3/50.2/\textbf{61.7}& &$\sigma_s$: 52.6/49.1/\textbf{56.6}&\\
		\hline
		
		\multirow{2}{*}{VGG+GPR~\cite{gao2019domain}}  &\multirow{2}{*}{VGG-16} 
		&$\sigma_l$: 56.3/61.0/52.2 & \multirow{2}{*}{105.8/\textbf{504.4}}  &$\sigma_l$: 52.5/55.8/49.6  &\multirow{2}{*}{127.3/\textbf{439.9}} \\
		\cline{3-3}	\cline{5-5}
		&&$\sigma_s$: 46.0/49.9/42.7& &$ \sigma_s$:42.6/45.3/40.2&\\
		\hline
		
		\multirow{2}{*}{RAZ\_Loc~\cite{liu2019recurrent}}  &\multirow{2}{*}{VGG-16} &$\sigma_l$: 62.5/69.2/56.9 & \multirow{2}{*}{128.7/665.4}  &$\sigma_l$: 59.8/66.6/54.3 &\multirow{2}{*}{151.5/634.7}   \\
		\cline{3-3}	\cline{5-5}
		&&$\sigma_s$: 54.5/60.5/49.6& &$\sigma_s$: 51.7/57.6/47.0&\\
		\hline
		
		\multirow{2}{*}{FPN* (\textbf{ours})} &\multirow{2}{*}{VGG-16}
		&$\sigma_l$: 64.4/60.0/\textbf{69.4} &\multirow{2}{*}{111.6/605.3}&$\sigma_l$: 58.9/65.9/53.3 &\multirow{2}{*}{143.2/603.2} \\
		\cline{3-3}	\cline{5-5}
		&&$\sigma_s$: 57.2/61.7/53.3&&$\sigma_s$: {51.0/57.1/46.1}&\\
		\hline
		
		\multirow{2}{*}{AutoScale* (\textbf{ours})}   &\multirow{2}{*}{VGG-16} &$\sigma_l$: \textbf{66.8}/70.1/63.8 &\multirow{2}{*}{\textbf{97.3}/571.2}&$\sigma_l$: \textbf{62.0}/67.3/57.4 &\multirow{2}{*}{\textbf{123.9}/515.5} \\
		\cline{3-3}	\cline{5-5}
		&&$\sigma_s$: \textbf{60.0}/62.9/57.3&&$\sigma_s$: \textbf{54.4}/59.1/50.4&\\
		\hline
	\end{tabular}
	\caption{Quantitative comparison of the localization performance on the NWPU-Crowd dataset in terms of precision (P), recall (R), and F-measure (F) using different adaptive distance tolerance thresholds $\sigma_l$ and $\sigma_s$ following~\cite{wang2020nwpu}.
	}
	\vspace{-8pt}
	\label{tab:nwpu_loc}
\end{table*}

\medskip
\noindent\textbf{Localization-based crowd counting result:}
We then evaluate the localization-based AutoScale using L2S. Some qualitative results in terms of distance-label maps are illustrated in Fig.~\ref{fig:locvis}. We also show some examples of person localization results in Fig.~\ref{fig:locationvis}. We present the bounding box on each location of people, generating from the KNN distance of each predicted local minima, which is similar as LSC-CNN~\cite{sam2020locate}. Comparied with LSC-CNN~\cite{sam2020locate}, AutoScale gives competitive bounding boxes and has better counting by localization performance in dense crowds. Qualitatively, the proposed distance-label map representation combined with the introduced dynamic cross-entropy loss is effective for localizing people even in dense regions. The proposed L2S effectively improves the localization and thus counting accuracy. The quantitative comparison with some other detection/localization-based methods is also depicted in Tab.~\ref{tab:NWPU_result}, Tab.~\ref{tab:JHU_result} and Tab.~\ref{tab:maemse} respectively. The localization-based AutoScale consistently outperforms state-of-the-art methods thanks to the L2S, the proposed localization method based on the local minima of distance-label map, and DCE loss on the ShanghaiTech Part A, UCF-QNRF, JHU-Crowd++ and NWPU-Crowd datasets. We have comparable performance with LSC-CNN~\cite{sam2020locate} on the ShanghaiTech Part B dataset.

To further demonstrate the effectiveness of the localization-based AutoScale, following~\cite{wang2020nwpu}, we evaluate the accuracy of localization using the common metrics in terms of precision, recall, and F-measure 
on the ShanghaiTech Part A, UCF-QNRF, and NWPU-Crowd dataset. The corresponding localization-based evaluation result is displayed in Tab.~\ref{tab:ShanghaiA_loc}, Tab.~\ref{tab:ucf_localization_metrics}, and Tab.~\ref{tab:nwpu_loc}, respectively.

For the ShanghaiTech Part A, as depicted in Tab.~\ref{tab:ShanghaiA_loc}, the proposed method improves the state-of-the-art method LSC-CNN~\cite{sam2020locate} by 22.6\% F-measure for the most strict setting $\sigma = 4$, and consistently improves upon the other methods for the less strict settings.

For the dense dataset UCF-QNRF, as shown in Tab.~\ref{tab:ucf_localization_metrics}, our localization-based AutoScale outperforms the state-of-the-art method LSC-CNN~\cite{sam2020locate} by 3.17 \% F-measure.

Finally, we compare the proposed method with some baselines such as~\cite{ren2015faster,hu2017finding,gao2019domain,liu2019recurrent} on the new released large-scale localization benchmark dataset NWPU-Crowd. As illustrated in Tab.~\ref{tab:nwpu_loc}, it can be observed that the proposed method largely improves the popular detection baseline Faster RCNN~\cite{ren2015faster} and TinyFaces~\cite{hu2017finding} by distinguishing margins in terms of both F-measure for localization and in terms of MAE/MSE for counting, even though they use a stronger backbone. Compared with the methods~\cite{gao2019domain,liu2019recurrent} dedicated for localization-based crowd counting, our localization-based AutoScale also outperforms them by at least 2.2 \% for $\sigma_l$ (2.7 \% for $\sigma_s$) F-measure.


\noindent\textbf{Counting and localization result on the extracted dense regions:} 
We also compare the performance of baseline and AutoScale on the extracted dense regions on the Shanghaitech Part A, Part B, UCF-QNRF, JHU-Crowd++, and NWPU-Crowd dataset. Since NWPU-Crowd does not release the ground-truth for the test set, we report results on the val set instead of test set for the other four datasets. As shown in Tab.~\ref{tab:mae_dense_region}, the proposed L2S achieves significant improvement compared with the baseline for both regression-based and localization-based counting on the extracted dense regions for all datasets.

\begin{table}[t]
\small
\centering
\setlength{\tabcolsep}{0.15mm}
\resizebox{0.47\textwidth}{!}{
\begin{tabular}{ |c|c|c|c|c|c|c|c|c|c|c|c|c|c|c| }
\hline
\multirow{2}{*}{Method}& \multicolumn{2}{c|}{Part A}& \multicolumn{2}{c|}{Part B} & \multicolumn{2}{c|}{QNRF} & \multicolumn{2}{c|}{JHU(test set)}& \multicolumn{2}{c|}{NWPU(val set)}\\
\cline{2-11}

&MAE &MSE &MAE &MSE &MAE & MSE &MAE &MSE &MAE &MSE\\
\hline
Baseline  & 94.2 & 137.1 & 12.1 &18.6 & 87.4&147.1& 76.9 &233.6 &81.3&446.3\\
AutoScale  & \textbf{71.5} & \textbf{91.0} &\textbf{9.3} & \textbf{12.3} & \textbf{78.0}&\textbf{121.5}&\textbf{67.9}&\textbf{190.3} &\textbf{70.7} &\textbf{329.6} \\
\hline
Baseline*  & 72.1 & 152.6 & 15.7 &28.7&138.5&271.0& 149.9&547.3 & 176.6 & 521.3\\
AutoScale*  & \textbf{55.1} & \textbf{95.4} & \textbf{8.5} & \textbf{13.3} & \textbf{110.5}&\textbf{190.9}&\textbf{109.6}&\textbf{436.8} &\textbf{125.2}&\textbf{314.1}\\

 \hline
\end{tabular}}
\caption{Quantitative comparison of baseline and AutoScale on the extracted dense regions.
}
\label{tab:mae_dense_region}
\end{table}

\subsection{Ablation study}
\label{subsec:ablation}

We conduct the ablation study mainly on the widely adopted ShanghaiTech Part A dataset. In the following, we study the effectiveness of the proposed L2S module, different designs in L2S, and the effectiveness of distance-label map with dynamic cross-entropy (DCE) loss.


\medskip
\noindent\textbf{Ablation study on effectiveness of the proposed L2S module.} We study the effectiveness of the proposed L2S from three aspects: 1) Effectiveness of L2S in alleviating the long-tailed distribution issue; 2) Effectiveness of applying L2S to different baseline methods; 3) Effectiveness of L2S compared with using fixed scale factors.

\begin{figure}[t]
	\begin{center}
	\includegraphics[width=1\linewidth]{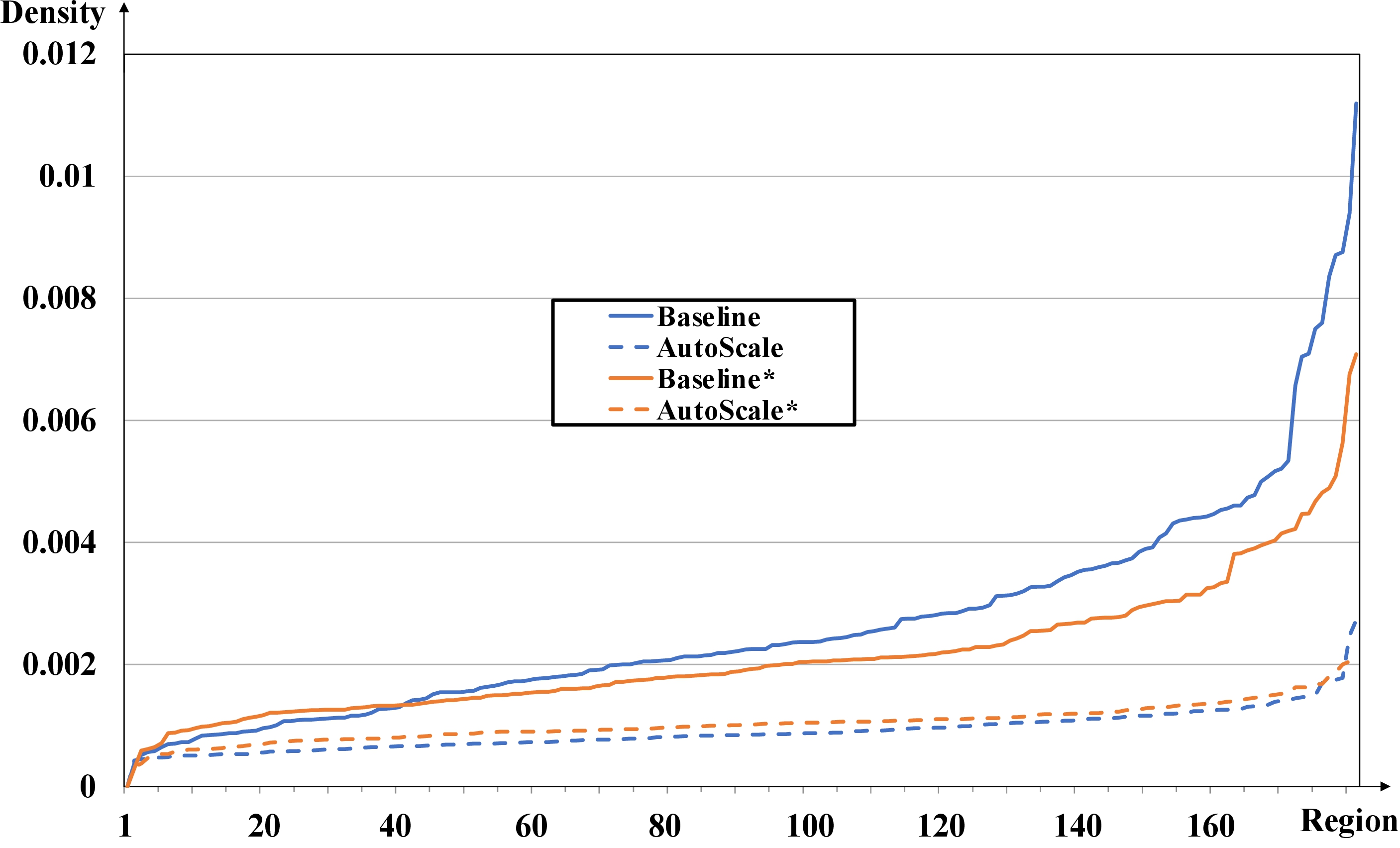}
	\end{center}
	\vspace{-12pt}
	\caption{Density of original selected dense regions and rescaled dense regions using L2S on the ShanghaiTech Part A dataset. $*$ stands for localization-based methods. We compute the density as the ratio between the total number of people in the region and the region size. The proposed L2S effectively brings all dense regions of different scales into similar density.}
	\label{fig:density_trend}
\end{figure}

\smallskip
\textit{Effectiveness of L2S in alleviating the long-tailed distribution issue:} We study the effect of alleviating the long-tailed distribution issue for both regression-based and localization-based AutoScale. Specifically,
for a given region $R$, we compute its density as the ratio between the total number of people in $R$ and the size of $R$. As depicted in Fig.~\ref{fig:density_trend}, for both regression and localization-based AutoScale, the automatically selected dense regions are of significantly varied density, presenting the lont-tailed distribution. The proposed L2S which normalizes the closeness effectively rescales all selected dense regions of different density into similar. Consequently, the appropriate scale factors lead pixel values to be decomposed and the overlapped blobs to be separated, meanwhile the similar closeness level and density makes the distribution of pixel values similar between sparse and dense regions, reducing the gaps between different regions in image and between different images and mitigating the long-tailed issue.

\smallskip
\textit{Effectiveness of applying L2S to different baseline methods:} Since L2S can improve the long-tailed distribution issue by rescaling the images, both training and inference phase thus benefit the model prediction and improve the accuracy. As shown in Tab.~\ref{tab:NWPU_result}, Tab.~\ref{tab:JHU_result}, and Tab.~\ref{tab:maemse}, the proposed L2S consistently improves the baseline FPN over a significant stage on all the datasets for both regression-based method and localization-based method.
Furthermore, as shown in Fig.~\ref{fig:regvis} and Fig.~\ref{fig:locvis}, the peak of each blob is more discriminative after the re-prediction by L2S. This is because the model is trained under a suitable pixel value/label distribution of ground truth, which helps the model to predict more accurately when an appropriate rescaled image is fed. 
Note that the resolution of the AutoScale density map is the same as the baseline density map. For visualization purpose, we simply resize the refined prediction for the rescaled dense region to the original size of the selected dense region, and replace the original density map on that dense region with such resized one.

Specifically, for the regression-based AutoScale, L2S improves the baseline model by 1.5/14.2 in MAE and 7.6/80.9 in MSE on the val/test set of NWPU-Crowd dataset, by 4.1/5.1 in MAE and 14.5/11.2 in MSE on the val/test set of JHU-Crowd++ dataset, by 4.5 in MAE and 11.3 in MSE on the UCF-QNRF dataset, by 5.1 in MAE and 11.5 in MSE on the ShanghaiTech Part A dataset. The improvement of L2S for localization-based AutoScale is more significant. Precisely, L2S improves the baseline model by 14.3/19.3 in MAE and 34.1/87.7 in MSE on the val/test set of NWPU-Crowd dataset, by 14.7/15.3 in MAE and 61.3/66.9 
in mse on the val/test set of JHU-Crowd++ dataset, by 20.4 in MAE and 60.5 in MSE on the UCF-QNRF dataset, by 9.9 in MAE and 38.3 in MSE on the ShanghaiTech Part A dataset. It is noteworthy L2S improves the baseline not as significantly as other datasets on the ShanghaiTech Part B since it is a relatively sparse dataset.

To demonstrate that the L2S can generalize to different models, we implement the MCNN~\cite{zhang2016single}, CSRNet~\cite{li2018csrnet}, CAN~\cite{liu2019context}, BL~\cite{ma2019bayesian}, and SFCN~\cite{wang2019learning} with the proposed L2S on the ShanghaiTech Part A dataset. The quantitative results are listed in Tab.~\ref{tab:regressor_l2s}, where we can observe that the proposed L2S is helpful to these baseline methods, consistently achieving noticeable improvements on the ShanghaiTech Part A dataset.

\begin{table}[!t]
\small
\footnotesize
\centering
\setlength{\tabcolsep}{1mm}

\begin{tabular}{ |l|c|c|c|c|c| }
\hline
Method &MAE&MSE\\
 \hline
MCNN$^*$~\cite{zhang2016single} &115.5& 174.1\\
MCNN*~\cite{zhang2016single} + L2S &\textbf{110.5}& \textbf{168.4}\\
\hline
CSRNET$^*$~\cite{li2018csrnet}  &69.2& 111.5\\
CSRNET$^*$~\cite{li2018csrnet} + L2S &\textbf{65.8}& \textbf{107.8}\\
\hline
BL$^*$~\cite{ma2019bayesian} &63.4& 102.9\\
BL$^*$~\cite{ma2019bayesian} + L2S &\textbf{60.9}& \textbf{100.7}\\
\hline
CAN$^*$~\cite{liu2019context} &66.5& 108.2\\
CAN$^*$~\cite{liu2019context} + L2S &\textbf{63.7}& \textbf{100.6}\\
\hline
SFCN$^*$ (VGG16)~\cite{wang2019learning} &68.3& 103.2\\
SFCN$^*$ (VGG16)~\cite{wang2019learning} + L2S &\textbf{64.1}& \textbf{100.9}\\
 \hline
\end{tabular}
\caption{
The effectiveness of L2S on different density regressors. $*$  means reproduction results using their corresponding officially released code.
}
\label{tab:regressor_l2s}
\end{table}

\begin{figure}[t]
	\begin{center}
	\includegraphics[width=1\linewidth]{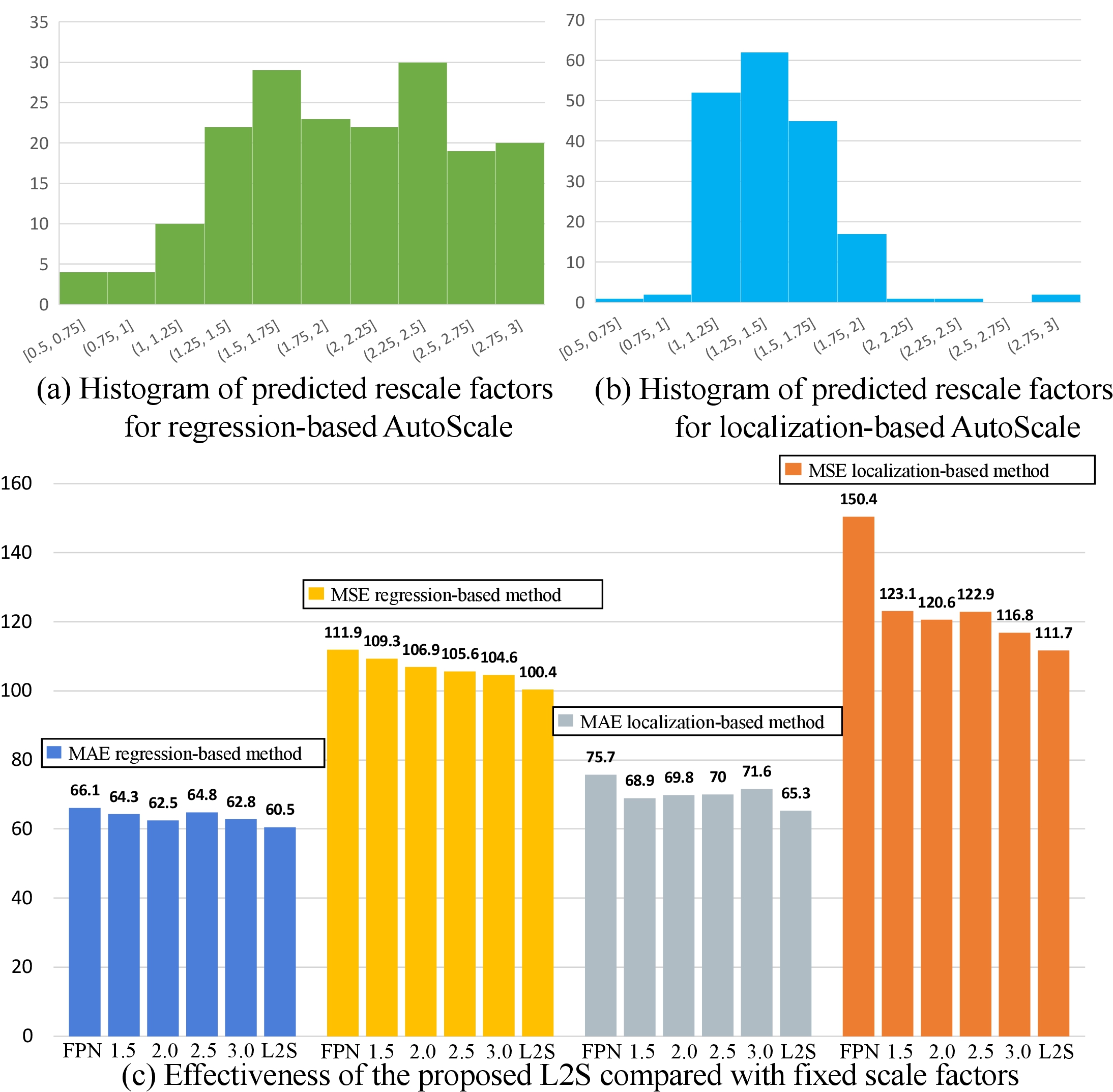}
	\end{center}
	\vspace{-12pt}
	\caption{The histogram of learned scale factors and the effectiveness of the proposed L2S compared with fixed scale factors.}
	\label{fig:histogram}
\end{figure}

\smallskip
\textit{Effectiveness of L2S compared with fixed scale factors:} To verify that the improvement is not simply given by zooming in dense regions for re-prediction, we conduct experiments by zooming in the selected dense regions at fixed scale factors. Based on the histogram of predicted rescale factors shown in Fig.~\ref{fig:histogram}(a) and (b) for both regression-based and localization-based AutoScale,
we fix the scale factors to 1.5, 2.0, 2.5, and 3. As shown in Fig.~\ref{fig:histogram} (c), though zooming in at fixed factors may improve the prediction on dense regions, the proposed L2S outperforms the alternatives using fixed scale factors, which is reasonable. In fact, since the selected regions are usually very dense, zooming in for re-prediction is beneficial for accurate counting, because it can relatively mitigate the long-tailed distribution. Nevertheless, since the selected dense regions are of different closeness levels (see Fig.~\ref{fig:density_trend}), we need adaptive scale factors for better counting. Indeed, L2S generates adaptive and appropriate scale factors (given by the ratio between the original closeness level and the re-scaled closeness level in Fig.~\ref{fig:density_trend}), which mitigates the long-tailed distribution in an adaptive and learnable manner and thus leads to more improvement of counting accuracy. We can also observe that in Fig.~\ref{fig:histogram} (a) and (b), there exist scale factors smaller than 1, which indicates that the center loss also forces the selected relatively sparse regions close to the central closeness level.

\medskip
\noindent\textbf{Ablation study on different designs in L2S.} We study the effect of some designs for L2S module including the density measure, the area-ratio threshold, and the ground-truth regeneration.

\smallskip
\textit{Effect of using different density measures:} We compare the performance of variants of AutoScale conducted with different density measures: the average people number and the average distance. As presented in Tab.~\ref{tab:new_density_measure}, the average-distance measure   consistently improves the performance over the average people-number, demonstrating the effectiveness of using the average distance as the density measure.

\begin{table}[t]
\small
\centering
\setlength{\tabcolsep}{1mm}
\resizebox{0.47\textwidth}{!}{
\begin{tabular}{ |l|c|c|c|c|c|c| }
\hline
\multirow{2}{*}{Method}& \multicolumn{2}{c|}{Part A}  & \multicolumn{2}{c|}{JHU++(test)}\\
\cline{2-5}
 &MAE &MSE & MAE &MSE \\
 \hline
FPN (baseline) & 66.1 &111.9 &81.5 & 303.9 \\
AutoScale (Average people number) & 64.5 & 106.6 &79.0&298.6\\
AutoScale (Average distance) & \textbf{60.5} & \textbf{100.4} &\textbf{76.4}&\textbf{292.7}\\
 \hline
\end{tabular}}
\caption{Performance comparison of using different density measures in the L2S module on the ShanghaiTech Part A and test set of JHU-Crowd++.}
\label{tab:new_density_measure}
\end{table}

\begin{figure}[t]
	\begin{center}
	\includegraphics[width=0.87\linewidth]{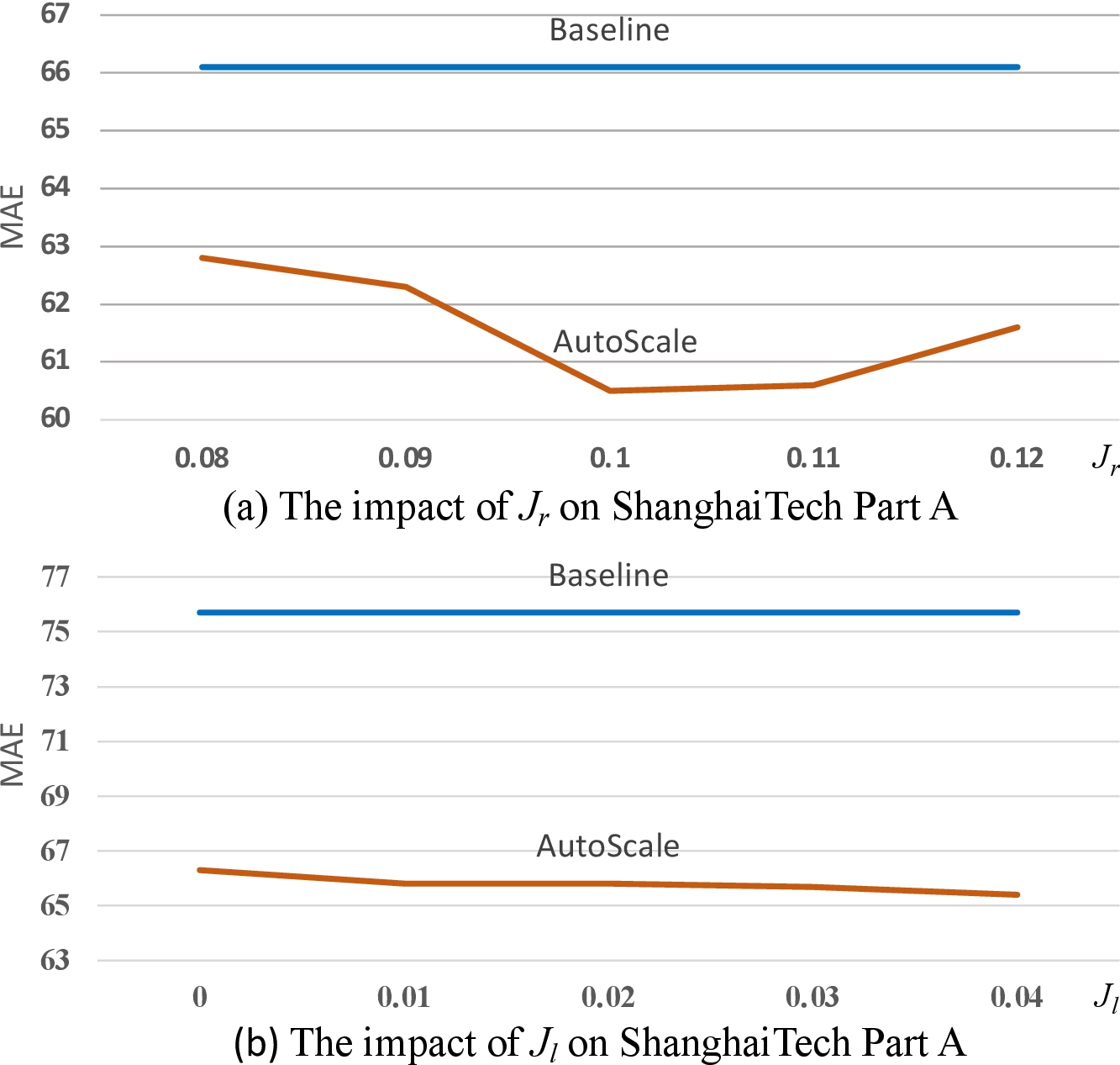}
	\end{center}
	\vspace{-12pt}
    \caption{The impact of different settings of $J_r$ and $J_l$ on the performance in terms of MAE.}
	\label{fig:effect_J}
\end{figure}

\smallskip
\textit{Effect of using different area-ratio threshold values:} We threshold the selected regions via $J_r$ (\textit{resp.} $J_l$) for the regression-based AutoScale (\textit{resp.} localization-based AutoScale). The performance changes with different $J_r$ and $J_l$ are presented in Fig.~\ref{fig:effect_J}. For very small values of $J_r$/$J_l$, we select to refine some very small regions which may be some noisy background regions, leading to slightly decreased performance. Using too large $J_r$/$J_l$ may ignore the refinement of some dense regions, which decreases a bit the performance. The good news is that the performance in terms of MAE is rather stable (varying one to three MAE) and the L2S consistently improves the baseline for a wide range of different values of $J_r$ and $J_l$. We roughly set $J_r$ to 0.1 and $J_l$ to 0.02 for all datasets and all experiments based on the experiments on the ShanghaiTech Part A dataset.

\begin{table}[t]
\small
\footnotesize
\centering
\setlength{\tabcolsep}{1mm}
\resizebox{0.47\textwidth}{!}{
\begin{tabular}{|c|c|c|c|c|c| }
    \hline
    Method &Gaussian kernel & Distance-label threshold & Size of dense region& MAE &MSE\\
    \hline
    \hline
    FPN (\textbf{baseline})&Fixed&-&-&66.1&111.9\\
    AutoScale & Multiplied &-&rescale&64.4&108.9\\
    AutoScale & Divided&- &keep& 65.1 & 109.4 \\
    AutoScale &Fixed &-&rescale&\textbf{60.5}&\textbf{100.4}\\
    \hline
    FPN* (\textbf{baseline})&- &Fixed &-&75.7 &150.4 \\
    AutoScale* &- & Multiplied &rescale& 72.9 & 129.1 \\
    AutoScale* &- &Divided &keep& 73.6& 133.3\\
    AutoScale*&-& Fixed&rescale&\textbf{65.8}&\textbf{112.1}\\
     \hline
    \end{tabular}
    }
\caption{Ablation study on kernel setting during training phase. ``Fixed'' means that the Gaussian kernel size/distance ranges for generating distance-label map are fixed even though the distances between Gaussian blobs /local minima are changed according to the learned scale factors, while ``Multiplied'' (\resp, ``Divided'') means that the Gaussian kernel size/distance ranges are multiplied (\resp, divided) by the learned scale factors. Note that when using ``Multiplied" (\resp, ``Divided"), we rescale (\resp, keep) the size of the selected dense region.}
\label{tab:Adaptive}
\end{table}

\smallskip
\textit{Effect of regenerating the ground-truth on extracted dense region:} To further demonstrate that the key to improve the performance is to mitigate the long-tailed distribution rather than simply zooming in the dense regions to an appropriate scale, we conduct an experiment that aims to compare fixed Gaussian kernels of ground-truth density map with scaled Gaussian kernels during the training phase. We perform two straightforward scaled Gaussian kernels: 1) rescale the size of selected region and multiply the kernel size with the scale factor; 2) keep the size of the selected region and divide the kernel size with the scale factor.
For the former variant, there still exist overlaps and pixel accumulations from different Gaussian blobs even though their centers are separated when the kernels becomes larger. In fact, rescaling both images and Gaussian blobs does not change the long-tailed distribution of the density values since the blobs are still overlapped the same relative amount. For the second variant, though using small kernels can mitigate the long-tailed distribution, it does not normalize the person size, which is beneficial to match the feature extractors of CNN. The adopted fixed Gaussian kernel with rescaled region effectively alleviates the long-tailed issue and normalizes the person size by rescaling all dense regions into similar and reasonable closeness level, leading to better performance. Indeed, as shown in Tab.~\ref{tab:Adaptive}, even though zooming in the images, the Scaled Gaussian kernel brings in some improvements, while the fixed Gaussian kernel significantly improves the baseline in terms of the density map regression. 
The same mechanism and observation also hold for the localization based on the distance-label map.

\begin{table}[t]
\small
\footnotesize
\centering
\setlength{\tabcolsep}{0.15mm}

\begin{tabular}{ |l|c|c|c|c|}
\hline
 Method& Density map & Distance-label map &MAE &  MSE \\
 \hline
FPN* & kernel = 8  & -&381.7 &528.6  \\
FPN* & kernel = 4 & -&282.9 &428.7\\
FPN* & kernel = 2 &-& 160.2 &289.9\\
FPN* & kernel = 1 &-& \textbf{109.5} &\textbf{213.2}\\
\hline
FPN* & - & Ordinal reg loss & 99.0 & 188.7 \\

FPN* & - &CE loss& 82.3 &159.8\\
FPN* & - & Focal loss & 79.3 & 160.2 \\
FPN* & - &DCE loss& 75.7 &150.4\\
AutoScale* & - &CE loss& 71.5 &127.6\\
AutoScale* & - & Focal loss & 70.1 & 122.9\\
AutoScale* & - &DCE loss& \textbf{65.8} &\textbf{112.1}\\

 \hline
\end{tabular}
\caption{Comparison of counting by localization based on classical density maps with different Gaussian kernels and the proposed distance label map combined with different loss functions. DCE loss denotes our proposed dynamic cross-entropy loss.
}
\label{tab:compdistdens}
\end{table}

\medskip
\noindent\textbf{Ablation study on effectiveness of the distance-label map with dynamic cross-entropy loss.} We study in the following the effect of the proposed distance-label map and the customized cross-entropy loss in localizing people.

\textit{The effectiveness of utilizing distance-label map for counting by localization:} To demonstrate the localization effectiveness of using the local minima of distance-label map representation, we compare with experiments using density map representation on the baseline model. We discard the L2S module for this ablation study. For the localization based on distance maps, the local minima represent exact person locations. For the density map representation, when a small spread parameter is used, there is few overlaps of Gaussian blobs between nearby people in dense regions. Therefore, person locations correspond to local maxima of density maps. Whereas, when a large spread parameter is used, the Gaussian blobs of nearby people in dense regions severely overlap each other, the maxima of such density map may not be accurate. 
Thus, we conduct experiments for localization with density maps using small spread parameters: \{1, 2, 4, 8\}. As depicted in Tab.~\ref{tab:compdistdens}, for localization using density map representation, a smaller spread parameter indeed yields a better count accuracy. Localization using the proposed distance-label map significantly outperforms localization based on density maps by 27.2 in MAE and 53.4 in MSE.

\textit{The effectiveness of utilizing the dynamic cross-entropy loss:}  
As shown in Tab.~\ref{tab:compdistdens}, the proposed dynamic cross-entropy loss significantly improves classical cross-entropy loss in predicting distance-label maps. Specifically, for the localization model without L2S (\textit{resp.} with L2S), the dynamic cross-entropy loss improves the results by 6.6 (\textit{resp.} 5.7) in MAE and 9.4 (\textit{resp.}, 15.5) in MSE. Moreover,
the proposed dynamic cross-entropy loss for the introduced distance-label map further boosts the localization accuracy, improving the localization using density maps by 33.8 in MAE and 62.8 in MSE.

 \begin{figure*}[t]
\centering
\includegraphics[width=0.75\paperwidth]{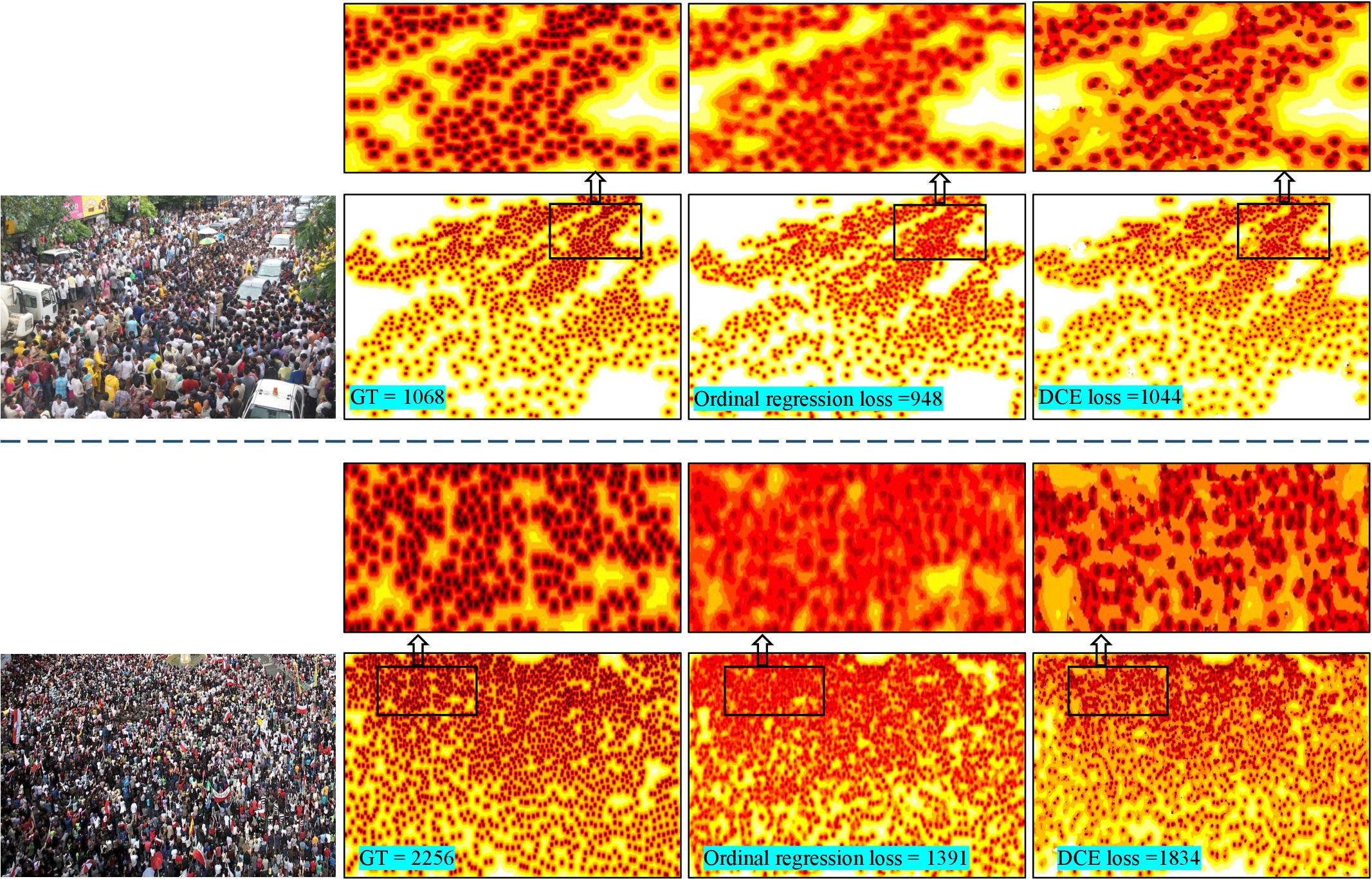}
\caption{Two examples of localization-based counting using the ordinal regression loss and the proposed DCE loss. }
\label{fig:ordinal_loss}
\end{figure*}

Besides, we also compare the proposed DCE loss with widely used focal loss~\cite{lin2017focal} and ordinal regression loss~\cite{fu2018deep}. Specifically, the proposed DCE loss outperforms the focal loss by 3.6 MAE (\textit{resp.} 4.3 MAE) and 9.8 MSE (\textit{resp.} 10.8 MSE) for the baseline  (\textit{resp.} AutoScale). The ordinal regression loss is dedicated for regressing ordinal numbers (\eg, $1, 2, 3, \dots$). We observe that the baseline with DCE loss achieves 23.3 MAE improvement over the ordinal regression loss. In fact, as shown in Fig.~\ref{fig:ordinal_loss}, though the ordinal regression loss guides the model to learn distance-label maps with similar patterns as that given by the proposed DCE loss. Yet, the local minima are not as accurate as the result given by the proposed DCE loss. A possible reason for this is that the regression loss usually results in blurred prediction, thus making it hard to localize the local minima.
\begin{table*}[t]
\centering
\setlength{\tabcolsep}{0.6mm}
\begin{tabular}{ |l|c
c|cc|cc|cc|cc|cc|cc| }
\hline
{\multirow{2}{*}{Method}}&\multicolumn{2}{c|}{A$\rightarrow$B}&\multicolumn{2}{c|}{B$\rightarrow$A}&\multicolumn{2}{c|}{QNRF$\rightarrow$A}&\multicolumn{2}{c|}{QNRF$\rightarrow$B}&\multicolumn{2}{c|}{A$\rightarrow$QNRF}&\multicolumn{2}{c|}{JHU$\rightarrow$A}&\multicolumn{2}{c|}{NWPU$\rightarrow$A}\\
\cline{2-15}
{} & MAE & MSE & MAE & MSE&MAE&MSE&MAE&MSE&MAE&MSE&MAE&MSE&MAE&MSE\\
\hline
\hline
MCNN~\cite{zhang2016single} (CVPR16)& 85.2 & 142.3 & 221.4&357.8 & -&-&- &-&- &-&-&- &-&- \\
D-ConvNet~\cite{shi2018crowd} (ECCV18) & 49.1 & 99.2 & 140.4&226.1&-&-& -&-&- &-&-&- &-&- \\
RRSP~\cite{wan2019residual} (CVPR19)& 40.0 & 68.5 & -&-&-&-& -&-&- &-&-&- &-&- \\
BL~\cite{ma2019bayesian} (ICCV19) &-&-&-&-&69.8&123.8&15.3&26.5&166.6 &287.6&- &-&-&-\\
\hline
FPN (\textbf{our}) & 32.3 &44.7 &143.6 &229.2& 73.0 &127.4 &13.2&22.5&164.0 &285.3&104.5&182.2&81.6&147.2\\
AutoScale (\textbf{ours}) & \textbf{30.9}&\textbf{42.6} &\textbf{130.8} &\textbf{220.9} &\textbf{69.6}&\textbf{121.0} &\textbf{12.0} &\textbf{21.2}&\textbf{138.1}&\textbf{231.9}&\textbf{101.8}&\textbf{186.2}&\textbf{78.2}&\textbf{137.0}\\
\hline
\hline
LSC-CNN*~\cite{sam2020locate} (TPAMI20) &\textbf{21.2}&\textbf{33.1}&150.2&244.6&97.0&154.6&\textbf{11.6}&\textbf{21.0}&198.5&359.1&-&-&-&-\\
\hline
FPN* (\textbf{ours}) & 23.0 &40.9 &133.6 &218.5& 76.8 &155.5 &15.8 &29.0&163.5 &301.6&125.0&226.9 &99.4&183.1\\
AutoScale* (\textbf{ours}) &23.1&39.0 &\textbf{131.3} &\textbf{210.3} & \textbf{71.9}&\textbf{129.9} &13.3 &22.2&\textbf{155.1}&\textbf{284.2}&\textbf{100.1} &\textbf{187.2}&\textbf{83.2}&\textbf{151.4}\\
\hline

\end{tabular}
\caption{Experimental results on the transferability of different methods under cross-dataset evaluation. The proposed L2S improves the transferability. * means localization-based methods.}
\label{tab:transfer}
\end{table*}

\subsection{Cross-dataset validation}
\label{subsec:crossdatasetval}
Since scene variation usually leads to significant performance drop, cross-dataset evaluation gradually attracts more attention in crowd counting~\cite{zhang2015cross,shi2018crowd,wan2019residual,ma2019bayesian}. In practice, a crowd counting method with strong generalizibility is usually expected. To verify the transferability of the proposed AutoScale, we evaluate the performance of the AutoScale under different cross-dataset validations. 
As depicted in Tab.~\ref{tab:transfer}, both the regression-based and localization-based AutoScale with L2S outperform state-of-the-art methods under cross-dataset evaluation, which demonstrates the superior generalizability of the proposed AutoScale. Specifically, the regression-based AutoScale with L2S outperforms RRSP~\cite{wan2019residual} (\textit{resp.} D-ConvNet~\cite{shi2018crowd}) by 9.1 (\textit{resp.} 18.2) in MAE and 25.9 (\textit{resp.} 56.6) in MSE on ShanghaiTech Part A crossing to Part B. Meanwhile, AutoScale with L2S outperforms BL~\cite{ma2019bayesian} consistently under the same cross-validation settings.

The localization-based AutoScale with L2S also improves the LSC-CNN~\cite{sam2020locate} by 18.9 in MAE and 34.3 in MSE on ShanghaiTech Part B crossing to Part A and by 25.1 in MAE and 24.7 in MSE on UCF-QNRF crossing to ShanghaiTech Part A. LSC-CNN is slightly better than us on the ShanghaiTech Part A crossing and UCF-QNRF crossing to ShanghaiTech Part B, because the proposed method L2S focuses more on the dense dataset. Moreover, we can see that the proposed L2S significantly improves the baseline models for both regression-based and localization-based counting, demonstrating the effectiveness of the proposed L2S under cross-dataset validation. 

\begin{figure*}[!t]
\centering
\includegraphics[width=0.75\paperwidth]{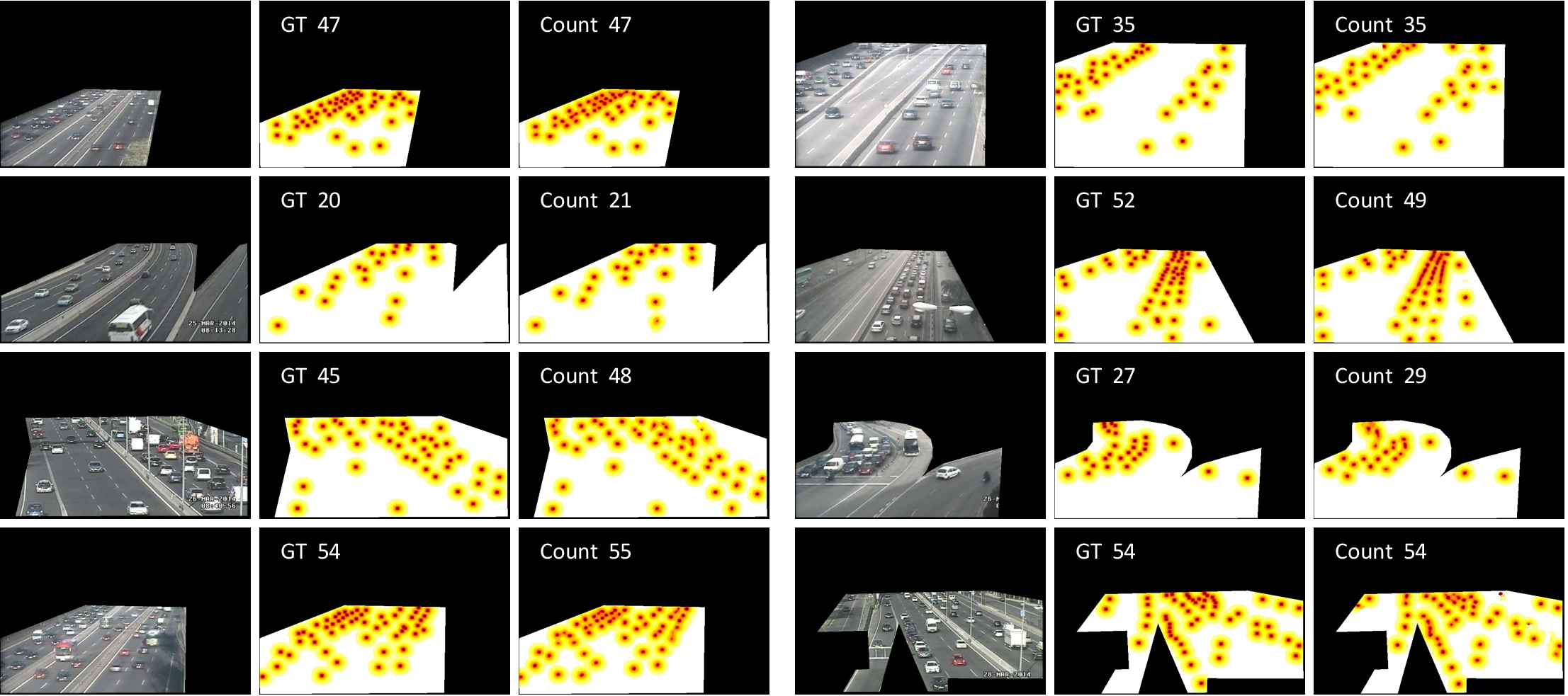}
\caption{Qualitative results in localizing vehicles using the proposed distance class label map and dynamic cross-entropy loss.}
\label{fig:vehicle}
\centering
\end{figure*}

\begin{table}[t]
\small
\footnotesize
\centering
\setlength{\tabcolsep}{1mm}
	\resizebox{0.47\textwidth}{!}{
\begin{tabular}{ |l|c|c|c|c|c| }
\hline
Method &GAME0&GAME1&GAME2&GAME3\\
 \hline
PSDDN*~\cite{liu2019point}(CVPR 19)&4.79& 5.43&6.68&8.40\\
LSC-CNN*~\cite{sam2020locate}(TPAMI 20) &4.60&5.40&6.90&\textbf{8.30}\\
  FPN* + CE Loss &3.22& 5.66 &7.42 &10.55\\
  FPN* + DCE Loss &\textbf{2.88 }&  \textbf{4.97 }&\textbf{6.64} &9.73\\
 \hline
\end{tabular}}
\caption{Quantitative comparison in counting vehicles on the TRANCOS dataset using the proposed distance-label map and dynamic cross-entropy loss. $*$ stands for localization-based methods.}
\label{tab:vehicle}
\end{table}

\subsection{Evaluation on vehicle counting}
\label{subsec:vehiclecount}
We also conduct an experiment on vehicle counting to further demonstrate the effectiveness of the distance-label map representation and the proposed dynamic cross-entropy loss other than on crowd counting. For that, we conduct experiments on the TRANCOS dataset~\cite{guerrero2015extremely}. We evaluate the proposed method using the associated metric Grid Average Mean Absolute Error ($GAME$) for this dataset, given by
\begin{equation}
    GAME(n)=\frac{1}{N_I}\sum_{i=1}^{N_I}(\sum_{j=1}^{4^n}|P_i^j-\hat{P}_i^j|),
\end{equation}
where $N_I$ is the number of images in the testing set, $P_i^j$ and $\hat{P}_i^j$ are the ground-truth and predicted count result of the $j$-th region in the $i$-th input image, and $n$
means that we evenly divide the whole image into $4^n$ non-overlapping regions. The $GAME(n)$ is  the sum of MAE in each of these non-overlapping regions. Note that $GAME(0)$ is equivalent to the $MAE$ metric in Eq.~\eqref{eq:metrics}.

Some qualitative results in terms of distance-label maps are illustrated in Fig.~\ref{fig:vehicle}. Using the baseline FPN model with distance-label maps and dynamic cross-entropy loss accurately localizes and counts the vehicles. The quantitative comparison with some state-of-the-art methods on this dataset is depicted in Tab.~\ref{tab:vehicle}. The proposed method outperforms the state-of-the-art methods~\cite{sam2020locate} under $GAME0$, $GAME1$, and $GAME2$ evaluations, and performs competitively with other methods under $GAME3$ evaluation. It is noteworthy to mention that the TRANCOS dataset provides regions of interest that mask the dense regions, therefore we only use the baseline model instead of the whole AutoScale to conduct the counting by localization experiment. This further confirms the effectiveness of the proposed distance-label map representation and dynamic cross-entropy loss in localizing vehicles other than people.

\begin{figure*}[t]
\centering
\includegraphics[width=0.75\paperwidth]{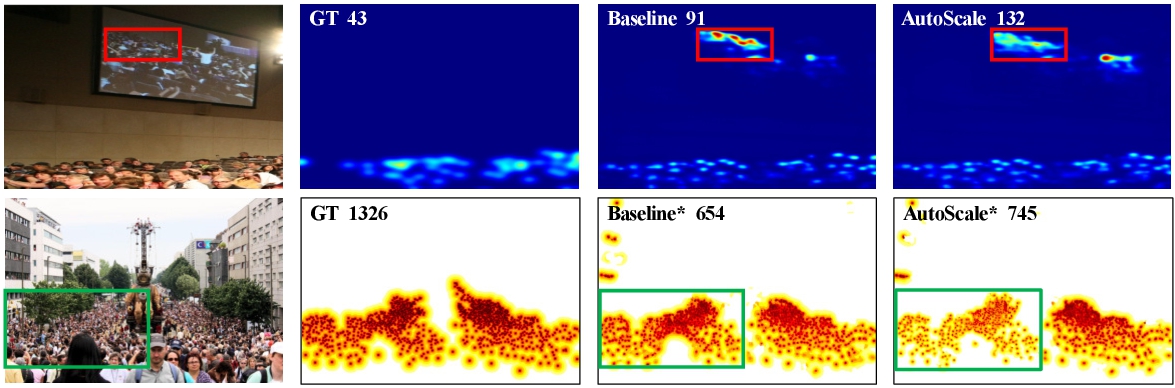}
\caption{Some failure cases of the regression-based and localization-based AutoScale. The enclosed regions are the automatically selected dense regions, which are rescaled via L2S for re-prediction.}
\label{fig:failure}
\centering
\end{figure*}

\subsection{Failure case and related discussion}
\label{subsec:failure}
Though both the regression and localization-based AutoScale improve the counting accuracy on dense regions in most cases. There are still some problems in very difficult images such as incorrect predictions on noisy backgrounds and very dense regions. Some failure cases are shown in Fig.~\ref{fig:failure}. In the first row, the initial prediction has difficulty in noisy background regions (containing ``fake'' people that do not count), leading to inappropriate dense region selection. Zooming in the noisy background region results in an uncontrollable count. 

\begin{table}[t]
\small
\footnotesize
\centering
\setlength{\tabcolsep}{1mm}
\resizebox{0.48\textwidth}{!}{
\begin{tabular}{ |c|c|c|c|c|c|c|c|c| }
\hline
\multirow{2}{*}{Method}& \multicolumn{2}{c|}{ShanghaiTech Part A}  & \multicolumn{2}{c|}{JHU-Crowd++ (test set)}\\
\cline{2-5}

&MAE &MSE &MAE &MSE \\
\hline
AutoScale-One  & 60.5 & 100.4 &76.4&292.7\\
AutoScale-Two  & 60.5 & 100.4 &\textbf{76.1}&\textbf{291.4} \\
\hline
AutoScale*-One & 65.8 & 112.1 &85.6&356.1\\
AutoScale*-Two & \textbf{65.6} & \textbf{109.3} &\textbf{84.9}&\textbf{354.6}\\
 \hline
\end{tabular}}
\caption{The effect of choosing at most one or two disconnected dense areas. * means our localization-based method.}
\label{tab:effect_multiple_regions}
\end{table}

Another case of unsatisfied counting is shown in the second row, where there are more than one very dense regions. Since the proposed AutoScale selects the largest dense region to rescale and refine the count, we still need to improve the count on the other dense regions. Therefore, it requires further exploration of the dense region selection methods. For that, we conduct experiments by simply choosing at most two disconnected dense areas for refinement by AutoScale. As depicted in Tab.~\ref{tab:effect_multiple_regions}, refining at most two dense regions slightly increases the performance on the ShanghaiTech Part A and JHU-Crowd++ dataset. Yet, the improvement is rather limited (less than 0.7 MAE and 2.8 MSE).

In practice, crowd people usually gather together, leading to very few images having more than one disconnected dense area (See Fig.~\ref{fig:regvis} and~\ref{fig:locvis} for example). Statistically, we find that none of ShanghaiTech Part A images and only 18 of 1600 images in JHU-Crowd++ test set have more than one disconnected dense area for the regression-based counting. For the localization-based counting, since we use a small $J_l = 0.02$, 45 of 182 images in ShanghaiTech Part A and 344 of 1600 images have more than one disconnected dense area. Yet, most of the rest candidate dense areas (except the largest one) are relatively sparse. This explains why the improvement is limited. Therefore, although it is true that choosing more dense regions for refinement makes more sense and brings slight improvement, we simply refine the prediction with L2S on the largest dense region. The currently adopted dense region selection strategy is still somehow heuristic, which could be improved and bring further performance gain in the future work.

\section{Conclusion}
\label{sec:conclusion}
In this paper, we explore how to learn and represent the density map better from the aspect of long-tailed distribution of pixel-wise density values for crowd counting. The proposed L2S performs in an unsupervised clustering way that leverages the center loss to bring all dense regions to similar and appropriate closeness levels for accurate counting. This is used for decomposing the accumulated density values and reducing the density distribution gap, helping to better learn the density maps. The effectiveness of L2S is validated on both density regression method and distance-label map localization method for crowd counting. For the later method, we also present a dedicated dynamic cross-entropy loss function to further improve the effectiveness of localizing people in addition to count people. Extensive experiments on five challenging datasets demonstrate the effectiveness of the proposed L2S module and the superior performance of the proposed localization-based counting in localizing people 
with the proposed dynamic cross-entropy loss. Besides, both the proposed density map regression AutoScale and localization-based AutoScale with L2S achieve appealing performance on the cross-dataset evaluation, showing a good transferability in crowd counting. Moreover, the proposed L2S can be applied to many other baseline methods, achieving consistent improvements.
Furthermore, the localization-based AutoScale also performs competitively with other methods in localizing and counting vehicles other than the crowd. 

In the future, we would like to improve the dense region selection strategy, which is currently somehow heuristic, and
continue to explore the issue of long-tailed distribution in the density map.
We are also interested in investigating the effectiveness of L2S in other vision tasks than the crowd or the vehicle counting and localization.

\section{Acknowledgement}
The work of Yongchao Xu was supported by the National Key Research and Development Program of China (2018AAA0100400), the National Natural Science Foundation of China (61936003 and 62176186), and in part by the Young Elite Scientists Sponsorship Program by CAST. The work of Xiang Bai was supported by the National Program for Support of Top-Notch Young Professionals and in part by the Program for HUST Academic Frontier Youth Team.

\bibliographystyle{spmpsci}      
\bibliography{reference.bib}

\begin{thebibliography}{100}
\providecommand{\url}[1]{{#1}}
\providecommand{\urlprefix}{URL }
\expandafter\ifx\csname urlstyle\endcsname\relax
  \providecommand{\doi}[1]{DOI~\discretionary{}{}{}#1}\else
  \providecommand{\doi}{DOI~\discretionary{}{}{}\begingroup
  \urlstyle{rm}\Url}\fi

\bibitem{arteta2014interactive}
Arteta, C., Lempitsky, V., Noble, J.A., Zisserman, A.: Interactive object
  counting.
\newblock In: Proc. of European Conference on Computer Vision, pp. 504--518.
  Springer (2014)

\bibitem{arteta2016counting}
Arteta, C., Lempitsky, V., Zisserman, A.: Counting in the wild.
\newblock In: Proc. of European Conference on Computer Vision, pp. 483--498.
  Springer (2016)

\bibitem{babu2018divide}
Babu~Sam, D., Sajjan, N.N., Venkatesh~Babu, R., Srinivasan, M.: Divide and
  grow: Capturing huge diversity in crowd images with incrementally growing
  cnn.
\newblock In: Proc. of IEEE Intl. Conf. on Computer Vision and Pattern
  Recognition, pp. 3618--3626 (2018)

\bibitem{badrinarayanan2017segnet}
Badrinarayanan, V., Kendall, A., Cipolla, R.: Segnet: A deep convolutional
  encoder-decoder architecture for image segmentation.
\newblock IEEE Transactions on Pattern Analysis and Machine Intelligence
  \textbf{39}(12), 2481--2495 (2017)

\bibitem{bai2020adaptive}
Bai, S., He, Z., Qiao, Y., Hu, H., Wu, W., Yan, J.: Adaptive dilated network
  with self-correction supervision for counting.
\newblock In: Proc. of IEEE Intl. Conf. on Computer Vision and Pattern
  Recognition, pp. 4594--4603 (2020)

\bibitem{baxes1994digital}
Baxes, G.A.: Digital image processing: principles and applications.
\newblock Wiley New York (1994)

\bibitem{brostow2006unsupervised}
Brostow, G.J., Cipolla, R.: Unsupervised bayesian detection of independent
  motion in crowds.
\newblock In: Proc. of IEEE Intl. Conf. on Computer Vision and Pattern
  Recognition, vol.~1, pp. 594--601 (2006)

\bibitem{cao2019learning}
Cao, K., Wei, C., Gaidon, A., Arechiga, N., Ma, T.: Learning imbalanced
  datasets with label-distribution-aware margin loss.
\newblock In: Advances in Neural Information Processing Systems, pp. 1565--1576
  (2019)

\bibitem{cao2018scale}
Cao, X., Wang, Z., Zhao, Y., Su, F.: Scale aggregation network for accurate and
  efficient crowd counting.
\newblock In: Proc. of European Conference on Computer Vision, pp. 734--750
  (2018)

\bibitem{chan2008privacy}
Chan, A.B., Liang, Z.S.J., Vasconcelos, N.: Privacy preserving crowd
  monitoring: Counting people without people models or tracking.
\newblock In: Proc. of IEEE Intl. Conf. on Computer Vision and Pattern
  Recognition, pp. 1--7 (2008)

\bibitem{chen2012feature}
Chen, K., Loy, C.C., Gong, S., Xiang, T.: Feature mining for localised crowd
  counting.
\newblock In: Proc. of BMVC, p.~3 (2012)

\bibitem{chen2010people}
Chen, T.Y., Chen, C.H., Wang, D.J., Kuo, Y.L.: A people counting system based
  on face-detection.
\newblock In: Proc. of Intl. Conf. on Genetic and Evolutionary Computing, pp.
  699--702 (2010)

\bibitem{cheng2019learning}
Cheng, Z.Q., Li, J.X., Dai, Q., Wu, X., Hauptmann, A.G.: Learning spatial
  awareness to improve crowd counting.
\newblock In: Porc. of IEEE Intl. Conf. on Computer Vision, pp. 6152--6161
  (2019)

\bibitem{cui2018large}
Cui, Y., Song, Y., Sun, C., Howard, A., Belongie, S.: Large scale fine-grained
  categorization and domain-specific transfer learning.
\newblock In: Proc. of IEEE Intl. Conf. on Computer Vision and Pattern
  Recognition, pp. 4109--4118 (2018)

\bibitem{dong2017class}
Dong, Q., Gong, S., Zhu, X.: Class rectification hard mining for imbalanced
  deep learning.
\newblock In: Porc. of IEEE Intl. Conf. on Computer Vision, pp. 1851--1860
  (2017)

\bibitem{fu2018deep}
Fu, H., Gong, M., Wang, C., Batmanghelich, K., Tao, D.: Deep ordinal regression
  network for monocular depth estimation.
\newblock In: Proceedings of the IEEE Conference on Computer Vision and Pattern
  Recognition, pp. 2002--2011 (2018)

\bibitem{gao2019domain}
Gao, J., Han, T., Wang, Q., Yuan, Y.: Domain-adaptive crowd counting via
  inter-domain features segregation and gaussian-prior reconstruction.
\newblock arXiv preprint arXiv:1912.03677  (2019)

\bibitem{gao2019c}
Gao, J., Lin, W., Zhao, B., Wang, D., Gao, C., Wen, J.: C\^{} 3 framework: An
  open-source pytorch code for crowd counting.
\newblock arXiv preprint arXiv:1907.02724  (2019)

\bibitem{ge2009marked}
Ge, W., Collins, R.T.: Marked point processes for crowd counting.
\newblock In: Proc. of IEEE Intl. Conf. on Computer Vision and Pattern
  Recognition, pp. 2913--2920 (2009)

\bibitem{geng2020recent}
Geng, C., Huang, S.j., Chen, S.: Recent advances in open set recognition: A
  survey.
\newblock IEEE Transactions on Pattern Analysis and Machine Intelligence
  (2020)

\bibitem{girshick2015fast}
Girshick, R.: Fast {R-CNN}.
\newblock In: Porc. of IEEE Intl. Conf. on Computer Vision, pp. 1440--1448
  (2015)

\bibitem{guerrero2015extremely}
Guerrero-G{\'o}mez-Olmedo, R., Torre-Jim{\'e}nez, B., L{\'o}pez-Sastre, R.,
  Maldonado-Basc{\'o}n, S., Onoro-Rubio, D.: Extremely overlapping vehicle
  counting.
\newblock In: Iberian Conference on Pattern Recognition and Image Analysis, pp.
  423--431. Springer (2015)

\bibitem{ha2016hypernetworks}
Ha, D., Dai, A., Le, Q.V.: Hypernetworks.
\newblock arXiv preprint arXiv:1609.09106  (2016)

\bibitem{he2009learning}
He, H., Garcia, E.A.: Learning from imbalanced data.
\newblock IEEE Transactions on knowledge and data engineering \textbf{21}(9),
  1263--1284 (2009)

\bibitem{he2017mask}
He, K., Gkioxari, G., Doll{\'a}r, P., Girshick, R.: Mask r-cnn.
\newblock In: Porc. of IEEE Intl. Conf. on Computer Vision, pp. 2961--2969
  (2017)

\bibitem{he2016deep}
He, K., Zhang, X., Ren, S., Sun, J.: Deep residual learning for image
  recognition.
\newblock In: Proc. of IEEE Intl. Conf. on Computer Vision and Pattern
  Recognition, pp. 770--778 (2016)

\bibitem{hossain2019crowd}
Hossain, M., Hosseinzadeh, M., Chanda, O., Wang, Y.: Crowd counting using
  scale-aware attention networks.
\newblock In: 2019 IEEE Winter Conference on Applications of Computer Vision
  (WACV), pp. 1280--1288. IEEE (2019)

\bibitem{hu2017finding}
Hu, P., Ramanan, D.: Finding tiny faces.
\newblock In: Proc. of IEEE Intl. Conf. on Computer Vision and Pattern
  Recognition, pp. 951--959 (2017)

\bibitem{hu2020count}
Hu, Y., Jiang, X., Liu, X., Zhang, B., Han, J., Cao, X., Doermann, D.:
  Nas-count: Counting-by-density with neural architecture search.
\newblock In: Proc. of European Conference on Computer Vision. Springer (2020)

\bibitem{huang2017densely}
Huang, G., Liu, Z., Van Der~Maaten, L., Weinberger, K.Q.: Densely connected
  convolutional networks.
\newblock In: Proc. of IEEE Intl. Conf. on Computer Vision and Pattern
  Recognition, pp. 4700--4708 (2017)

\bibitem{idrees2015detecting}
Idrees, H., Soomro, K., Shah, M.: Detecting humans in dense crowds using
  locally-consistent scale prior and global occlusion reasoning.
\newblock EEE Trans. Pattern Anal. Mach. Intell. \textbf{37}(10), 1986--1998
  (2015)

\bibitem{idrees2018composition}
Idrees, H., Tayyab, M., Athrey, K., Zhang, D., Al-Maadeed, S., Rajpoot, N.,
  Shah, M.: Composition loss for counting, density map estimation and
  localization in dense crowds.
\newblock In: Proc. of European Conference on Computer Vision. Springer (2018)

\bibitem{jaderberg2015spatial}
Jaderberg, M., Simonyan, K., Zisserman, A., et~al.: Spatial transformer
  networks.
\newblock In: Proc. of Advances in Neural Information Processing Systems, pp.
  2017--2025 (2015)

\bibitem{jiang2019mask}
Jiang, S., Lu, X., Lei, Y., Liu, L.: Mask-aware networks for crowd counting.
\newblock IEEE Transactions on Circuits and Systems for Video Technology
  (2019)

\bibitem{jiang2019crowd}
Jiang, X., Xiao, Z., Zhang, B., Zhen, X., Cao, X., Doermann, D., Shao, L.:
  Crowd counting and density estimation by trellis encoder-decoder networks.
\newblock In: Proc. of IEEE Intl. Conf. on Computer Vision and Pattern
  Recognition, pp. 6133--6142 (2019)

\bibitem{jiang2019learning}
Jiang, X., Zhang, L., Lv, P., Guo, Y., Zhu, R., Li, Y., Pang, Y., Li, X., Zhou,
  B., Xu, M.: Learning multi-level density maps for crowd counting.
\newblock IEEE transactions on neural networks and learning systems  (2019)

\bibitem{jiang2020attention}
Jiang, X., Zhang, L., Xu, M., Zhang, T., Lv, P., Zhou, B., Yang, X., Pang, Y.:
  Attention scaling for crowd counting.
\newblock In: Proc. of IEEE Intl. Conf. on Computer Vision and Pattern
  Recognition, pp. 4706--4715 (2020)

\bibitem{kang2018crowd}
Kang, D., Chan, A.: Crowd counting by adaptively fusing predictions from an
  image pyramid.
\newblock Proc. of BMVC  (2018)

\bibitem{kang2018beyond}
Kang, D., Ma, Z., Chan, A.B.: Beyond counting: Comparisons of density maps for
  crowd analysis tasks—counting, detection, and tracking.
\newblock IEEE Transactions on Circuits and Systems for Video Technology
  \textbf{29}(5), 1408--1422 (2018)

\bibitem{kingma2014adam}
Kingma, D.P., Ba, J.: Adam: A method for stochastic optimization.
\newblock In: Proc. of International Conference on Learning Representations
  (2014)

\bibitem{laradji2018blobs}
Laradji, I.H., Rostamzadeh, N., Pinheiro, P.O., Vazquez, D., Schmidt, M.: Where
  are the blobs: Counting by localization with point supervision.
\newblock In: Proc. of European Conference on Computer Vision, pp. 547--562
  (2018)

\bibitem{li2018csrnet}
Li, Y., Zhang, X., Chen, D.: {CSRNet}: Dilated convolutional neural networks
  for understanding the highly congested scenes.
\newblock In: Proc. of IEEE Intl. Conf. on Computer Vision and Pattern
  Recognition, pp. 1091--1100 (2018)

\bibitem{lian2019density}
Lian, D., Li, J., Zheng, J., Luo, W., Gao, S.: Density map regression guided
  detection network for {RGB-D} crowd counting and localization.
\newblock In: Proc. of IEEE Intl. Conf. on Computer Vision and Pattern
  Recognition, pp. 1821--1830 (2019)

\bibitem{lin2017feature}
Lin, T.Y., Doll{\'a}r, P., Girshick, R.B., He, K., Hariharan, B., Belongie,
  S.J.: Feature pyramid networks for object detection.
\newblock In: Proc. of IEEE Intl. Conf. on Computer Vision and Pattern
  Recognition, vol.~1, p.~4 (2017)

\bibitem{lin2017focal}
Lin, T.Y., Goyal, P., Girshick, R., He, K., Doll{\'a}r, P.: Focal loss for
  dense object detection.
\newblock In: Porc. of IEEE Intl. Conf. on Computer Vision, pp. 2980--2988
  (2017)

\bibitem{liu2015bayesian}
Liu, B., Vasconcelos, N.: Bayesian model adaptation for crowd counts.
\newblock In: Porc. of IEEE Intl. Conf. on Comp. Vis., pp. 4175--4183 (2015)

\bibitem{liu2019recurrent}
Liu, C., Weng, X., Mu, Y.: Recurrent attentive zooming for joint crowd counting
  and precise localization.
\newblock In: Proc. of IEEE Intl. Conf. on Computer Vision and Pattern
  Recognition, pp. 1217--1226 (2019)

\bibitem{liu2018decidenet}
Liu, J., Gao, C., Meng, D., Hauptmann, A.G.: Decidenet: counting varying
  density crowds through attention guided detection and density estimation.
\newblock In: Proc. of IEEE Intl. Conf. on Computer Vision and Pattern
  Recognition, pp. 5197--5206 (2018)

\bibitem{liu2020WeighingCounts}
Liu, L., Lu, H., Zou, H., Xiong, H., Cao, Z., Chun, H.: Weighing counts:
  Sequential crowd counting by reinforcement learning (2020)

\bibitem{liu2019crowd}
Liu, L., Qiu, Z., Li, G., Liu, S., Ouyang, W., Lin, L.: Crowd counting with
  deep structured scale integration network.
\newblock In: Porc. of IEEE Intl. Conf. on Computer Vision, pp. 1774--1783
  (2019)

\bibitem{liu2018crowd}
Liu, L., Wang, H., Li, G., Ouyang, W., Lin, L.: Crowd counting using deep
  recurrent spatial-aware network.
\newblock IJCAI  (2018)

\bibitem{liu2019adcrowdnet}
Liu, N., Long, Y., Zou, C., Niu, Q., Pan, L., Wu, H.: Adcrowdnet: An
  attention-injective deformable convolutional network for crowd understanding.
\newblock In: Proc. of IEEE Intl. Conf. on Computer Vision and Pattern
  Recognition, pp. 3225--3234 (2019)

\bibitem{liu2019context}
Liu, W., Salzmann, M., Fua, P.: Context-aware crowd counting.
\newblock In: Proc. of IEEE Intl. Conf. on Computer Vision and Pattern
  Recognition, pp. 5099--5108 (2019)

\bibitem{liu2019exploiting}
Liu, X., Van De~Weijer, J., Bagdanov, A.D.: Exploiting unlabeled data in cnns
  by self-supervised learning to rank.
\newblock IEEE Transactions on Pattern Analysis and Machine Intelligence
  (2019)

\bibitem{liu2018leveraging}
Liu, X., van~de Weijer, J., Bagdanov, A.D.: Leveraging unlabeled data for crowd
  counting by learning to rank.
\newblock In: Proc. of IEEE Intl. Conf. on Computer Vision and Pattern
  Recognition (2018)

\bibitem{liu2020adaptive}
Liu, X., Yang, J., Ding, W.: Adaptive mixture regression network with local
  counting map for crowd counting.
\newblock In: Proc. of European Conference on Computer Vision. Springer (2020)

\bibitem{liu2019point}
Liu, Y., Shi, M., Zhao, Q., Wang, X.: Point in, box out: Beyond counting
  persons in crowds.
\newblock In: Proc. of IEEE Intl. Conf. on Computer Vision and Pattern
  Recognition, pp. 6469--6478 (2019)

\bibitem{luo2020hybrid}
Luo, A., Yang, F., Li, X., Nie, D., Jiao, Z., Zhou, S., Cheng, H.: Hybrid graph
  neural networks for crowd counting.
\newblock Proc. of the AAAI Conf. on Artificial Intelligence  (2020)

\bibitem{ma2019bayesian}
Ma, Z., Wei, X., Hong, X., Gong, Y.: Bayesian loss for crowd count estimation
  with point supervision.
\newblock In: Porc. of IEEE Intl. Conf. on Computer Vision, pp. 6142--6151
  (2019)

\bibitem{miaoshallow}
Miao, Y., Lin, Z., Ding, G., Han, J.: Shallow feature based dense attention
  network for crowd counting.
\newblock Proc. of the AAAI Conf. on Artificial Intelligence  (2020)

\bibitem{najibi2018autofocus}
Najibi, M., Singh, B., Davis, L.S.: Autofocus: Efficient multi-scale inference.
\newblock Porc. of IEEE Intl. Conf. on Computer Vision  (2019)

\bibitem{oh2019crowd}
Oh, M.h., Olsen, P.A., Ramamurthy, K.N.: Crowd counting with decomposed
  uncertainty.
\newblock Proc. of the AAAI Conf. on Artificial Intelligence  (2020)

\bibitem{oh2016deep}
Oh~Song, H., Xiang, Y., Jegelka, S., Savarese, S.: Deep metric learning via
  lifted structured feature embedding.
\newblock In: Proc. of IEEE Intl. Conf. on Computer Vision and Pattern
  Recognition, pp. 4004--4012 (2016)

\bibitem{olmschenk2019improving}
Olmschenk, G., Tang, H., Zhu, Z.: Improving dense crowd counting convolutional
  neural networks using inverse k-nearest neighbor maps and multiscale
  upsampling.
\newblock arXiv preprint arXiv:1902.05379  (2019)

\bibitem{onoro2016towards}
Onoro-Rubio, D., L{\'o}pez-Sastre, R.J.: Towards perspective-free object
  counting with deep learning.
\newblock In: Proc. of European Conference on Computer Vision, pp. 615--629
  (2016)

\bibitem{ouyang2016factors}
Ouyang, W., Wang, X., Zhang, C., Yang, X.: Factors in finetuning deep model for
  object detection with long-tail distribution.
\newblock In: Proc. of IEEE Intl. Conf. on Computer Vision and Pattern
  Recognition, pp. 864--873 (2016)

\bibitem{ranjan2018iterative}
Ranjan, V., Le, H., Hoai, M.: Iterative crowd counting.
\newblock In: Proc. of European Conference on Computer Vision (2018)

\bibitem{recasens2018learning}
Recasens, A., Kellnhofer, P., Stent, S., Matusik, W., Torralba, A.: Learning to
  zoom: a saliency-based sampling layer for neural networks.
\newblock In: Proc. of European Conference on Computer Vision, pp. 51--66
  (2018)

\bibitem{ren2015faster}
Ren, S., He, K., Girshick, R., Sun, J.: Faster r-cnn: Towards real-time object
  detection with region proposal networks.
\newblock In: Proc. of Advances in Neural Information Processing Systems, pp.
  91--99 (2015)

\bibitem{ribera2019}
Ribera, J., G\"{u}era, D., Chen, Y., Delp, E.J.: Locating objects without
  bounding boxes.
\newblock Proc. of IEEE Intl. Conf. on Computer Vision and Pattern Recognition
  (2019).
\newblock {Long Beach, CA}

\bibitem{rodriguez2011density}
Rodriguez, M., Laptev, I., Sivic, J., Audibert, J.Y.: Density-aware person
  detection and tracking in crowds.
\newblock In: Porc. of IEEE Intl. Conf. on Computer Vision, pp. 2423--2430
  (2011)

\bibitem{sajid2020zoomcount}
Sajid, U., Sajid, H., Wang, H., Wang, G.: Zoomcount: A zooming mechanism for
  crowd counting in static images.
\newblock IEEE Transactions on Circuits and Systems for Video Technology
  (2020)

\bibitem{salakhutdinov2011learning}
Salakhutdinov, R., Torralba, A., Tenenbaum, J.: Learning to share visual
  appearance for multiclass object detection.
\newblock In: CVPR 2011, pp. 1481--1488. IEEE (2011)

\bibitem{sam2020locate}
Sam, D.B., Peri, S.V., Sundararaman, M.N., Kamath, A., Radhakrishnan, V.B.:
  Locate, size and count: Accurately resolving people in dense crowds via
  detection.
\newblock IEEE Transactions on Pattern Analysis and Machine Intelligence
  (2020)

\bibitem{sam2017switching}
Sam, D.B., Surya, S., Babu, R.V.: Switching convolutional neural network for
  crowd counting.
\newblock In: Proc. of IEEE Intl. Conf. on Computer Vision and Pattern
  Recognition, vol.~1, p.~6 (2017)

\bibitem{shi2019revisiting}
Shi, M., Yang, Z., Xu, C., Chen, Q.: Revisiting perspective information for
  efficient crowd counting.
\newblock In: Proc. of IEEE Intl. Conf. on Computer Vision and Pattern
  Recognition, pp. 7279--7288 (2019)

\bibitem{shi2019counting}
Shi, Z., Mettes, P., Snoek, C.G.: Counting with focus for free.
\newblock In: Porc. of IEEE Intl. Conf. on Computer Vision, pp. 4200--4209
  (2019)

\bibitem{shi2018crowd}
Shi, Z., Zhang, L., Liu, Y., Cao, X., Ye, Y., Cheng, M.M., Zheng, G.: Crowd
  counting with deep negative correlation learning.
\newblock In: Proc. of IEEE Intl. Conf. on Computer Vision and Pattern
  Recognition, pp. 5382--5390 (2018)

\bibitem{vgg16network}
Simonyan, K., Zisserman, A.: Very deep convolutional networks for large-scale
  image recognition.
\newblock In: Proc. of International Conference on Learning Representations
  (2015)

\bibitem{sindagi2017cnn}
Sindagi, V.A., Patel, V.M.: Cnn-based cascaded multi-task learning of
  high-level prior and density estimation for crowd counting.
\newblock In: Proc. of IEEE Intl. Conf. on Advanced Video and Signal Based
  Surveillance, pp. 1--6 (2017)

\bibitem{sindagi2017generating}
Sindagi, V.A., Patel, V.M.: Generating high-quality crowd density maps using
  contextual pyramid cnns.
\newblock In: Porc. of IEEE Intl. Conf. on Computer Vision (2017)

\bibitem{sindagi2018survey}
Sindagi, V.A., Patel, V.M.: A survey of recent advances in cnn-based single
  image crowd counting and density estimation.
\newblock Pattern Recognition Letters \textbf{107}, 3--16 (2018)

\bibitem{sindagi2019ha}
Sindagi, V.A., Patel, V.M.: {HA-CCN}: Hierarchical attention-based crowd
  counting network.
\newblock IEEE Transactions on Image Processing  (2019).
\newblock Accepted

\bibitem{sindagi2019multi}
Sindagi, V.A., Patel, V.M.: Multi-level bottom-top and top-bottom feature
  fusion for crowd counting.
\newblock In: Porc. of IEEE Intl. Conf. on Computer Vision, pp. 1002--1012
  (2019)

\bibitem{sindagi2019pushing}
Sindagi, V.A., Yasarla, R., Patel, V.M.: Pushing the frontiers of unconstrained
  crowd counting: New dataset and benchmark method.
\newblock In: Porc. of IEEE Intl. Conf. on Computer Vision, pp. 1221--1231
  (2019)

\bibitem{sindagi2020jhu}
Sindagi, V.A., Yasarla, R., Patel, V.M.: Jhu-crowd++: Large-scale crowd
  counting dataset and a benchmark method.
\newblock IEEE Transactions on Pattern Analysis and Machine Intelligence
  (2020)

\bibitem{singh2018analysis}
Singh, B., Davis, L.S.: An analysis of scale invariance in object
  detection--snip.
\newblock In: Proc. of IEEE Intl. Conf. on Computer Vision and Pattern
  Recognition, pp. 3578--3587 (2018)

\bibitem{singh2018sniper}
Singh, B., Najibi, M., Davis, L.S.: Sniper: Efficient multi-scale training.
\newblock In: Proc. of Advances in Neural Information Processing Systems, pp.
  9310--9320 (2018)

\bibitem{tian2019padnet}
Tian, Y., Lei, Y., Zhang, J., Wang, J.Z.: Padnet: Pan-density crowd counting.
\newblock IEEE Transactions on Image Processing  (2019)

\bibitem{van2017devil}
Van~Horn, G., Perona, P.: The devil is in the tails: Fine-grained
  classification in the wild.
\newblock arXiv preprint arXiv:1709.01450  (2017)

\bibitem{viola2005detecting}
Viola, P., Jones, M.J., Snow, D.: Detecting pedestrians using patterns of
  motion and appearance.
\newblock International Journal of Computer Vision \textbf{63}(2), 153--161
  (2005)

\bibitem{wan2019adaptive}
Wan, J., Chan, A.: Adaptive density map generation for crowd counting.
\newblock In: Porc. of IEEE Intl. Conf. on Computer Vision, pp. 1130--1139
  (2019)

\bibitem{wan2019residual}
Wan, J., Luo, W., Wu, B., Chan, A.B., Liu, W.: Residual regression with
  semantic prior for crowd counting.
\newblock In: Proc. of IEEE Intl. Conf. on Computer Vision and Pattern
  Recognition, pp. 4036--4045 (2019)

\bibitem{wang2011automatic}
Wang, M., Wang, X.: Automatic adaptation of a generic pedestrian detector to a
  specific traffic scene.
\newblock In: Proc. of IEEE Intl. Conf. on Computer Vision and Pattern
  Recognition, pp. 3401--3408 (2011)

\bibitem{wang2020nwpu}
Wang, Q., Gao, J., Lin, W., Li, X.: Nwpu-crowd: A large-scale benchmark for
  crowd counting and localization.
\newblock IEEE Transactions on Pattern Analysis and Machine Intelligence
  (2020).
\newblock \doi{10.1109/TPAMI.2020.3013269}

\bibitem{wang2019learning}
Wang, Q., Gao, J., Lin, W., Yuan, Y.: Learning from synthetic data for crowd
  counting in the wild.
\newblock In: Proc. of IEEE Intl. Conf. on Computer Vision and Pattern
  Recognition, pp. 8198--8207 (2019)

\bibitem{wang2017learning}
Wang, Y.X., Ramanan, D., Hebert, M.: Learning to model the tail.
\newblock In: Advances in Neural Information Processing Systems, pp. 7029--7039
  (2017)

\bibitem{xiong2019open}
Xiong, H., Lu, H., Liu, C., Liu, L., Cao, Z., Shen, C.: From open set to closed
  set: Counting objects by spatial divide-and-conquer.
\newblock In: Porc. of IEEE Intl. Conf. on Computer Vision, pp. 8362--8371
  (2019)

\bibitem{xu2019learn}
Xu, C., Qiu, K., Fu, J., Bai, S., Xu, Y., Bai, X.: Learn to scale: Generating
  multipolar normalized density map for crowd counting.
\newblock Porc. of IEEE Intl. Conf. on Computer Vision  (2019)

\bibitem{yan2019perspective}
Yan, Z., Yuan, Y., Zuo, W., Tan, X., Wang, Y., Wen, S., Ding, E.:
  Perspective-guided convolution networks for crowd counting.
\newblock In: Porc. of IEEE Intl. Conf. on Computer Vision, pp. 952--961 (2019)

\bibitem{yang2020reverse}
Yang, Y., Li, G., Wu, Z., Su, L., Huang, Q., Sebe, N.: Reverse perspective
  network for perspective-aware object counting.
\newblock In: Proc. of IEEE Intl. Conf. on Computer Vision and Pattern
  Recognition, pp. 4374--4383 (2020)

\bibitem{zhang2019relational}
Zhang, A., Shen, J., Xiao, Z., Zhu, F., Zhen, X., Cao, X., Shao, L.: Relational
  attention network for crowd counting.
\newblock In: Porc. of IEEE Intl. Conf. on Computer Vision, pp. 6788--6797
  (2019)

\bibitem{zhang2019attentional}
Zhang, A., Yue, L., Shen, J., Zhu, F., Zhen, X., Cao, X., Shao, L.: Attentional
  neural fields for crowd counting.
\newblock In: Porc. of IEEE Intl. Conf. on Computer Vision, pp. 5714--5723
  (2019)

\bibitem{zhang2015cross}
Zhang, C., Li, H., Wang, X., Yang, X.: Cross-scene crowd counting via deep
  convolutional neural networks.
\newblock In: Proc. of IEEE Intl. Conf. on Computer Vision and Pattern
  Recognition, pp. 833--841 (2015)

\bibitem{zhang2019nonlinear}
Zhang, L., Shi, Z., Cheng, M.M., Liu, Y., Bian, J.W., Zhou, J.T., Zheng, G.,
  Zeng, Z.: Nonlinear regression via deep negative correlation learning.
\newblock EEE Trans. Pattern Anal. Mach. Intell.  (2019).
\newblock Accepted

\bibitem{zhang2019wide}
Zhang, Q., Chan, A.B.: Wide-area crowd counting via ground-plane density maps
  and multi-view fusion cnns.
\newblock In: Proc. of IEEE Intl. Conf. on Computer Vision and Pattern
  Recognition, pp. 8297--8306 (2019)

\bibitem{zhang20203d}
Zhang, Q., Chan, A.B.: 3d crowd counting via multi-view fusion with 3d gaussian
  kernels.
\newblock Proc. of the AAAI Conf. on Artificial Intelligence  (2020)

\bibitem{zhang2017range}
Zhang, X., Fang, Z., Wen, Y., Li, Z., Qiao, Y.: Range loss for deep face
  recognition with long-tailed training data.
\newblock In: Porc. of IEEE Intl. Conf. on Computer Vision, pp. 5409--5418
  (2017)

\bibitem{zhang2016single}
Zhang, Y., Zhou, D., Chen, S., Gao, S., Ma, Y.: Single-image crowd counting via
  multi-column convolutional neural network.
\newblock In: Proc. of IEEE Conf. on Comp. Vis. and Patt. Rec., pp. 589--597
  (2016)

\bibitem{zhao2019leveraging}
Zhao, M., Zhang, J., Zhang, C., Zhang, W.: Leveraging heterogeneous auxiliary
  tasks to assist crowd counting.
\newblock In: Proc. of IEEE Intl. Conf. on Computer Vision and Pattern
  Recognition, pp. 12736--12745 (2019)

\bibitem{zhao2009people}
Zhao, X., Delleandrea, E., Chen, L.: A people counting system based on face
  detection and tracking in a video.
\newblock In: Proc. of IEEE Intl. Conf. on Advanced Video and Signal Based
  Surveillance, pp. 67--72 (2009)

\bibitem{zhao2016crossing}
Zhao, Z., Li, H., Zhao, R., Wang, X.: Crossing-line crowd counting with
  two-phase deep neural networks.
\newblock In: Proc. of European Conference on Computer Vision, pp. 712--726.
  Springer (2016)

\bibitem{zheng2017learning}
Zheng, H., Fu, J., Mei, T., Luo, J.: Learning multi-attention convolutional
  neural network for fine-grained image recognition.
\newblock In: Porc. of IEEE Intl. Conf. on Computer Vision, pp. 5209--5217
  (2017)

\bibitem{zhu2014capturing}
Zhu, X., Anguelov, D., Ramanan, D.: Capturing long-tail distributions of object
  subcategories.
\newblock In: Proc. of IEEE Intl. Conf. on Computer Vision and Pattern
  Recognition, pp. 915--922 (2014)

\bibitem{zhu2016we}
Zhu, X., Vondrick, C., Fowlkes, C.C., Ramanan, D.: Do we need more training
  data?
\newblock International Journal of Computer Vision \textbf{119}(1), 76--92
  (2016)

\end{thebibliography}

\end{document}